\documentclass[Afour,sageh,times]{sagej}

\usepackage{moreverb,url}
\usepackage{graphicx}
\usepackage{subcaption}
\usepackage{xcolor}
\usepackage{mdframed}
\usepackage{multirow}
\usepackage{algorithm}
\usepackage{algpseudocode}
\usepackage{tcolorbox}
\usepackage{enumitem}

   \newcommand{\ie}{\textit{i.e.}}
   \newcommand{\eg}{\textit{e.g.}}

\usepackage[colorlinks,bookmarksopen,bookmarksnumbered,citecolor=blue,urlcolor=blue]{hyperref}

\newcommand\BibTeX{{\rmfamily B\kern-.05em \textsc{i\kern-.025em b}\kern-.08em
T\kern-.1667em\lower.7ex\hbox{E}\kern-.125emX}}
\def\volumeyear{2025}
\newcommand{\tl}{\boldsymbol{t}_\text{left}} 
\newcommand{\tr}{\boldsymbol{t}_\text{right}} 
\newcommand{\A}{\mathcal{A}}  
\newcommand{\Kc}{\boldsymbol{K}_c} 
\newcommand{\Kzero}{\boldsymbol{K}_0} 
\newcommand{\K}{\boldsymbol{K}} 
\newcommand{\pzero}{p_{i,0}} 
\newcommand{\szero}{s_{i,0}} 
\newcommand{\La}{\mathcal{L}} 

\begin{document}

\runninghead{Deng et al.}

\title{CLASP: General-Purpose Clothes Manipulation with Semantic Keypoints}

\author{Yuhong Deng\affilnum{1}, Chao Tang\affilnum{1,2}, Cunjun Yu\affilnum{1}, Linfeng Li\affilnum{1} and David Hsu\affilnum{1}}

\affiliation{\affilnum{1}School of Computing, Smart System Institute, National University of Singapore, Singapore\\
\affilnum{2}Department of Electronic and Electrical Engineering, Southern University of Science and Technology, China}

\corrauth{David Hsu, National University of Singapore, Singapore}

\email{dyhsu@comp.nus.edu.sg}

\begin{abstract}
Clothes manipulation, such as folding or hanging, is a critical capability for home service robots. Despite recent advances, most existing methods remain limited to specific clothes types and tasks, due to the complex, high-dimensional geometry of clothes. This paper presents \textit{CLothes mAnipulation with Semantic keyPoints} (CLASP), which aims at \textit{general-purpose} clothes manipulation over diverse clothes types, T-shirts, shorts, skirts, long dresses, \ldots\,, as well as different tasks, folding, flattening, hanging, \ldots\,. The core idea of CLASP is \textit{semantic keypoints}---e.g., ``left sleeve'' and ``right shoulder''---a sparse spatial-semantic representation, salient for both perception and action. Semantic keypoints of clothes can be reliably extracted from RGB-D images and provide an effective representation for a wide range of clothes manipulation policies. CLASP uses semantic keypoints as an intermediate representation to connect high-level task planning and low-level action execution. At the high level, it exploits vision language models (VLMs) to predict task plans over the semantic keypoints. At the low level, it executes the plans with the help of a set of pre-built manipulation skills conditioned on the keypoints.  Extensive simulation experiments show that CLASP outperforms state-of-the-art baseline methods on multiple tasks across diverse clothes types, demonstrating strong performance and generalization. Further experiments with a Franka dual-arm system on four distinct tasks---folding, flattening, hanging, and placing---confirm CLASP's performance on real-life clothes manipulation. The CLASP source code is available online at \hyperlink{https://github.com/dengyh16code/CLASP}{https://github.com/dengyh16code/CLASP}.
\end{abstract}

\keywords{Deformable object manipulation, state representation, foundation models}

\maketitle

\section{Introduction}

People have long anticipated intelligent service robots performing home laundry chores: fold or hang shirts, pants, gowns \ldots\,. While there has been significant progress in robot clothes manipulation in recent years~\citep{speedfolding, mesh_dynamics, flatten_1, particle_dynamics},  existing methods are usually tailored to specific clothes and tasks. \textit{General-purpose} clothes manipulation---a single robot system performing different manipulation tasks over diverse clothes types---remains an important open challenge.  

A core issue here is representation. 
Clothes are deformable objects with high-dimensional state space and complex geometry that varies significantly across categories, \eg, shirts and gowns. How do we represent clothes?
Dense physics-based representations, such as particles~\citep{particle_2, particle_1}, can model diverse clothes, but the fine details required are difficult to estimate from visual observations, because of clothes deformation and self-occlusion~\citep{survey_deformable}. Sparse keypoints~\citep{landmark,keypoint_synthetic} are more robust to deformation, but generalize poorly over different clothes~\citep{cloth_survey}. Both approaches focus exclusively on geometry, while ignoring semantics. Semantics is crucial for clothes representation, as clothes are man-made objects \textit{designed specifically for humans}.

\begin{figure*}[ht]
    \centering
    \vspace{0.3cm}
    \includegraphics[width=\linewidth]{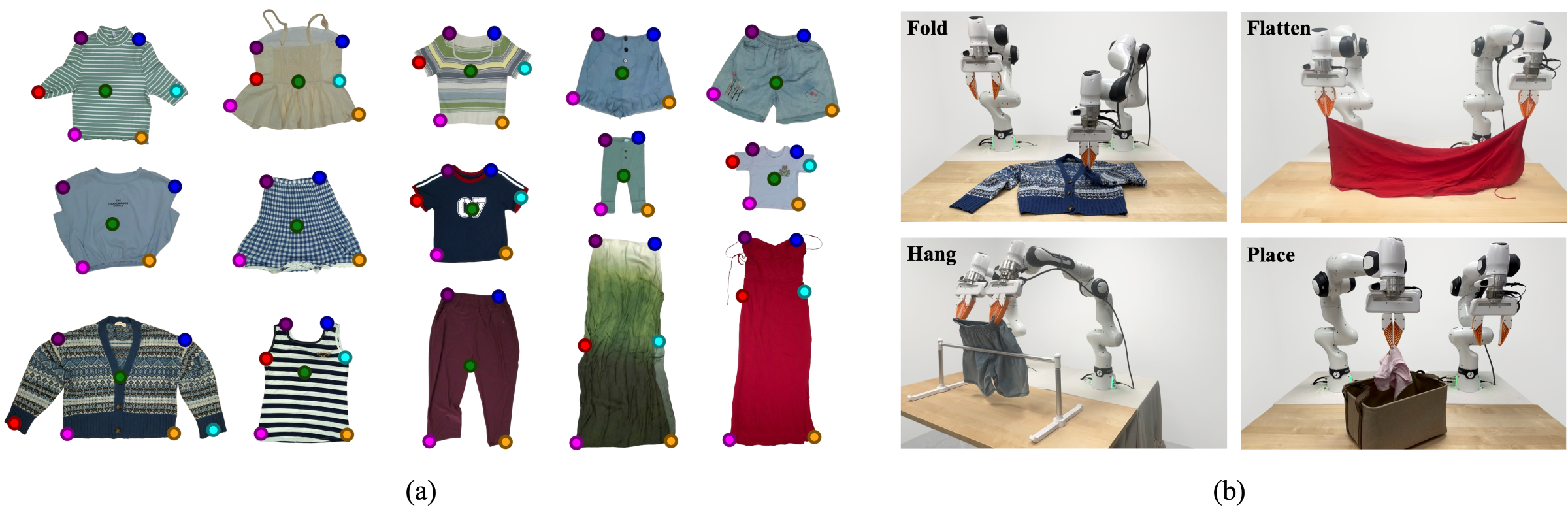}
    \caption{CLothes mAnipulation with Semantic keyPoints (CLASP). The semantic keypoint representation enables CLASP to generalize over many different types of clothes and tasks. (a) Semantic keypoints for various types of clothes. (b) Four distinct clothes manipulation tasks in our experiments. }
     \label{fig:tasks}
\end{figure*}

We introduce \textit{semantic keypoints} (Figure~\ref{fig:tasks}), \eg, ``left sleeve'' and ``right shoulder'', as a spatial–semantic representation that connects perception and action for general-purpose clothes manipulation. Semantic keypoints provide significant advantages. First, they are salient for both perception and action.  They can be extracted reliably with RGB-D sensors under moderate deformation and occlusion. They also identify commonly manipulated regions on clothes and highlight their affordance,  providing an abstraction to guide action planning and execution. Second, semantic keypoint descriptors naturally emerge in human language due to their salience.  Finally, semantic keypoints are grounded to structural features consistent within the same clothes category, enabling effective generalization.

Semantic keypoint extraction poses two main challenges: (i) identifying semantically meaningful keypoints for open-category clothes, and (ii) achieving accurate spatial localization of keypoints under deformation and occlusion. To decouple the semantic and spatial challenges, we propose a two-stage method.  First, we use a vision-language model (VLM) for semantic understanding of individual clothes categories. For each category, we determine the semantic keypoints on a \textit{prototype} clothes instance lying flat. Next, we match these keypoints spatially to novel input clothes instances under possible deformation and occlusion, using vision foundation models. 

Given a natural language instruction and RGB-D observations, \textit{CLothes mAnipulation with Semantic keyPoints} (CLASP) performs various manipulation tasks over a wide range of clothes types. To tackle the challenge of general-purpose clothes manipulation, CLASP leverages semantic keypoints to decompose the manipulation task into a two-level hierarchy (Figure~\ref{fig:method}). At the high level,  CLASP uses a VLM  to transform the input language instruction into a task plan over the semantic keypoints. For example,  grasp the left sleeve, move to the center, \ldots\,. It verifies the generated task plan for geometric feasibility through motion planning and decides whether to replan. At the low level, CLASP executes each subtask in the high-level plan, using a set of pre-built manipulation skills conditioned on the keypoints. After executing each subtask, CLASP closes the loop by updating the visual observation and replanning if needed. The process continues until the task is successfully completed.
To evaluate CLASP, we conducted experiments both in simulation and on a dual-arm robot system. For simulation experiments, we extended the SoftGym~\citep{softgym} benchmark and experimented with four clothes categories and four representative tasks under various scenarios. CLASP outperforms well-established baselines in clothes manipulation, including FlingBot~\citep{flingbot} and FabricFlownet~\citep{fabricflownet}. For real-robot experiments, we implemented CLASP on Franka robots for dual-arm manipulation on 15 clothes of various types. The results confirm the performance of CLASP and its generality. They demonstrate that semantic keypoints provide an effective representation of clothes manipulation, enabling general-purpose clothes manipulation with foundation models.\par

 This work extends our earlier paper~\citep{clasp_icra} in three main aspects:
 \begin{itemize}
     \item  \textit{Semantic keypoint extraction for open-category clothes}. We improved semantic keypoint extraction by leveraging foundation models, eliminating the need for labeled training data (Section~\ref{sec:keypoint_extraction}). This significantly enhances the generalization performance of CLASP on open-category clothes.
     \item \textit{Closed-loop  planning}.  CLASP now closes the loop at both the task planning and the subtask execution levels (Section~\ref{sec:task_planning}). Closed-loop planning substantially improves overall system performance (Sec.~\ref{subsec:task_plan}), especially with crumpled clothes.
     \item \textit{Expanded real-world evaluation}. We added diverse clothes, such as T-shirts, sweaters, shorts, pants, skirts, and dresses, and conducted comprehensive real-robot experiments to validate the effectiveness and generalization of CLASP (Section~\ref{subsec:real_expri}).
 \end{itemize}

The rest of this paper is organized as follows. The related work is reviewed in Section~\ref{sec:relatedwork}. Section~\ref{sec:overview} introduces the overall CLASP framework. Section~\ref{sec:keypoint_extraction} presents semantic keypoints extraction. Section~\ref{sec:clothes_manipulation} details clothes manipulation guided by semantic keypoints. The experimental setup and results are provided in Section~\ref{sec:experiments}.
Section~\ref{sec:discussion} discusses limitations and future directions. Finally, Section~\ref{sec:conclusion} concludes the paper. Additional implementation and experimental details are provided in the appendix.
\section{Related Work}
\label{sec:relatedwork}

\subsection{State Representation for Deformable Object} 
  Effective state representation is crucial for object manipulation. However, representing the state of deformable objects, such as clothes, remains a significant challenge due to their high-dimensional state space~\citep{survey_deformable,cloth_survey}. Some task-specific state representations have been developed, such as using an approximate polygonal representation for clothes folding~\citep{fold_3} and topological coordinates for assistive dressing~\citep{dressing}. However, designing a general representation across different objects and tasks remains a significant challenge. \par
  Physics-based models such as particle~\citep{particle_2, particle_1, particle-general-1} and mesh~\citep{mesh_2, mesh_1} provide a solution to model deformable dynamics and represent a wide range of objects. However, these dense models make accurate state estimation of deformable objects challenging, especially under deformation and occlusion~\citep{mesh_3}.\par 
  Recently, learning to encode raw observations into a compact latent representation has emerged as a more effective solution~\citep{latent_space_1, latent_learning_model}. However, they often struggle to generalize to new environments due to distribution shifts caused by variations in visual conditions~\citep{foresight_fabric}.\par
  Keypoint representation provides a succinct way for deformable object representation, enabling efficient planning and learning~\citep{keypoint_policy, keypoint_graph, unigarment} while remaining robust to visual condition variations. Previous keypoints representation focuses on geometric information, using dense keypoints to represent the topology of deformable objects~\citep{skeleton,landmark,clothesnet}. However, keypoint extraction is still challenged by occlusion and deformation, while dense keypoints often lack consistency and generalization. More recently, \cite{rekep} utilize vision foundation models to extract sparse keypoints and employs a vision-language model to impose keypoint-centric constraints for clothes manipulation, though it requires domain-specific human intervention.
  
  In this paper, we present semantic keypoints as a general spatial-semantic representation of clothes. Semantic keypoints carry semantic meaning and can be described in natural language, making them sparse and salient for perception and action. Semantic keypoints are easy to extract, generalize well, and indicate where clothes are more likely to be manipulated. 

\subsection{Deformable Object Manipulation}
Existing deformable manipulation approaches can be categorized into model-based and model-free approaches.
Model-based approaches use a dynamic model to predict configurations of deformable objects under a certain action and then select the appropriate action~\citep{model-base2, fold_3, model-based}. The key challenge lies in the dynamic modeling of deformable objects. To tackle this, two main techniques are used: physics-based models~\citep{particle_3, particle-general-2}, which rely on physical law, and data-driven models~\citep{learning_dynamic_1,mesh_3}, which are learned from interaction data. Typical physics-based models include particle-based models~\citep{particle_dynamics, particle-general-1} and mesh-based models~\citep{mesh_dynamics}. Recently, Material Point Method (MPM)~\citep{MPM} has emerged as a new physics-based modeling method, which combines particles with grid-based forces for more accurate dynamic modeling~\citep{maniskill2}. However, physics-based modes often struggle with heavy parameter tuning~\citep{differentiable_2}. Data-driven dynamics models are learned from interaction data, typically formulated as a supervised learning problem~\citep{model-base1, learning_dynamic_1,latent_learning_model}. Compared to physics-based models, data-driven models support a more flexible state space, especially via latent representation~\citep{latent_space_1,deformnet}. However, data-driven models rely on expensive robot interaction data and often fail to generalize to different objects and tasks. Overall, the generalization of model-based approaches highly depends on accurate dynamic modeling and calibration for unseen objects and tasks~\citep{cloth_survey}, which remains an open problem due to the complex, nonlinear dynamics of deformable objects.\par 

Model-free approaches aim to learn robot actions directly without establishing a dynamic model, which can be categorized as imitation learning and reinforcement learning methods. The crucial challenge of imitation learning methods is collecting expert demonstrations, which can be gathered from human~\citep{cloth_video_1,cloth_video} or scripted policies~\citep{script_1, fabricflownet, fold_2}. Specifically, \cite{speedfolding} employ imitation learning to acquire quasi-static action primitives, such as pick-and-place and dragging, for bimanual cloth folding. \cite{cloth_video} propose a method to learn fabric smoothing and folding policies from human demonstration videos. To avoid collecting extensive expert demonstrations, reinforcement learning methods offer an alternative solution. \cite{hang_1} develop a sim-to-real reinforcement learning method for towel hanging. \cite{flingbot} use reinforcement learning to learn a dynamic fling policy for efficient cloth flattening. \par
Although model-free approaches have made significant progress in a wide range of tasks~\citep{learning_clothes, flatten_2, bag_1}, the generalization remains a significant limitation. Most of the previous methods aim to learn task-specific policies~\citep{single_policy, fabricflownet,bag_2}. Although some goal-conditioned methods use goal images for multi-task learning~\citep{goal_conditioned_2, goal_conditioned_1}, they often struggle to generalize to unseen goals. To enhance generalization, collecting more data~\citep{unifolding} and domain randomization~\citep{domain_random_1,dressing} have been explored. However, model-free approaches still struggle to generalize to unseen tasks or object categories. In this paper, we aim to tackle general-purpose clothes manipulation across diverse clothes categories and task types. To this end, we propose CLASP, a framework that integrates semantic keypoint representations with foundation models. CLASP enables generalization to a wide range of clothes and manipulation tasks. 

 \begin{figure*}[ht]
    \centering
    \vspace{0.3cm}
    \includegraphics[width=\linewidth]{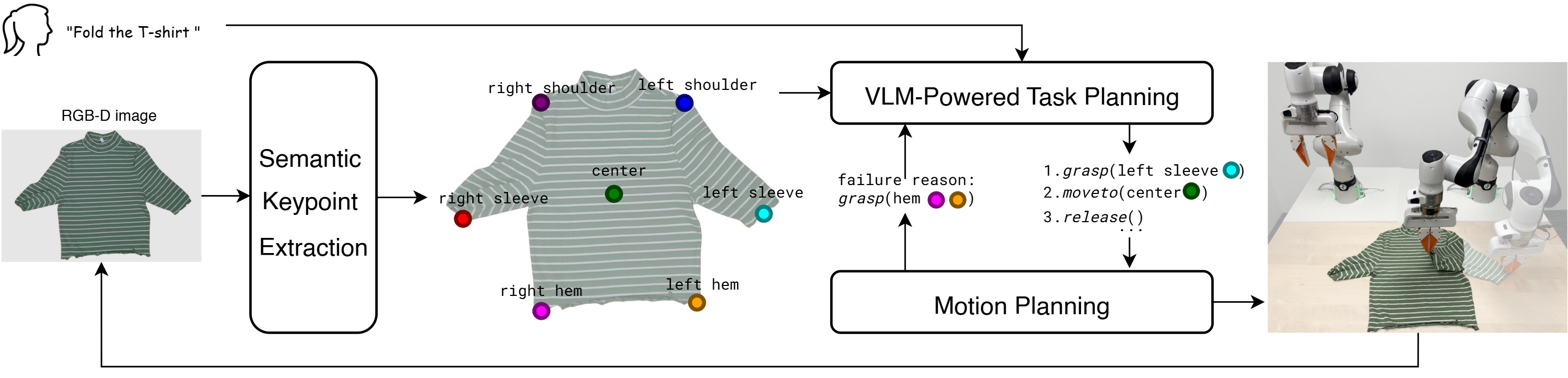}
    \caption{CLASP Overview. Given an RGB-D observation, CLASP extracts semantic keypoints. These keypoints, along with the RGB image and task instruction, are fed to a VLM to generate a task plan. Once the plan is verified for feasibility through motion planning, the subtasks are executed sequentially. After each subtask execution, CLASP updates the observation and decides whether to replan. This process repeats until the overall task is completed.}
    \label{fig:method}
\end{figure*}

\subsection{Language-Conditioned Manipulation}
Language-conditioned manipulation is an emerging field that aims to enable seamless interaction between humans and robots by teaching robots to understand and execute instructions specified in natural language~\citep{languge_conditioned_2,language_conditioned_0,language_manipulation_survey}. The recent emergence of large foundation models that are pretrained on internet-scale data, such as LLMs and VLMs, has significantly advanced language-conditioned manipulation tasks. These models provide commonsense knowledge for reasoning~\citep{llm_manipulation_1,video-language-model} and perception~\citep{llm_manipulation_3, physical_vlm}, significantly enhancing generalization capabilities. More recently, vision–language–action models (VLAs)~\citep{openvla, pi_0,fast} fine-tune VLMs for robot control, aiming for generalist policies but relying on large-scale robotic data.\par
Beyond specifying tasks, natural language also serves as a powerful abstraction of affordance~\citep{grounding_language}, capturing the functions of an object, \ie, how it can be manipulated. For task-oriented grasping~\citep{task-oriented}, GraspGPT~\citep{graspgpt} utilizes object class descriptions generated by a large language model (LLM) to identify optimal grasping regions. Similarly, ~\cite{tool_affordance} provide natural language descriptions of tools to accelerate learning language-conditioned tool manipulation policies. RoboPoint finetunes a VLM for spatial affordance prediction conditioned on natural language~\citep{robopoint}. MimicFunc~\citep{mimicfunc} leverages functional cues to identify keypoints for tool manipulation.\par

Most previous language-conditioned object manipulation methods focus on rigid objects, with few addressing deformable object manipulation~\citep{language_def, bifold}.  For rigid objects, existing methods are available to execute actions such as picking, placing, moving, and grasping~\citep{llm_manipulation_2}. However, a significant gap remains between task planning and action execution when it comes to deformable object manipulation. Motivated by prior work that uses natural language as an abstraction of affordance, we propose semantic keypoints as a spatial-semantic representation of clothes. These keypoints act as an interface between VLM-powered task planning and low-level action execution, enabling general-purpose clothes manipulation.
\section{Overview}
\label{sec:overview}
For general-purpose clothes manipulation, CLASP leverages semantic keypoints to decompose the task into a two-level hierarchy. Figure~\ref{fig:method} illustrates the overall framework. Given RGB-D observations of clothes and a language instruction, CLASP generates a sequence of manipulation actions to complete the specified task.\par

First, CLASP extracts semantic keypoints using foundation models. Each extracted keypoint is represented by a 3D position and an associated language descriptor. These keypoints, together with the RGB image and task instruction, are then fed into a vision-language model (VLM) to generate a task plan over semantic keypoints (\eg, grasp(left sleeve), moveto(center), ...). Each subtask in the plan is grounded by selecting the appropriate skill from a pre-built skill library. The plan is then verified for feasibility through motion planning, which generates the corresponding robot trajectories for execution. After each subtask, CLASP updates its observation and determines whether replanning is required. This closed-loop process continues until the task is successfully completed.\par

In the following sections, we will introduce how we extract effective semantic keypoints of open-category clothes in Sec.~\ref{sec:keypoint_extraction}. Then, to enable general-purpose clothes manipulation with semantic keypoints, we present the construction of the skill library in Sec.~\ref{sec:basic_skill}, followed by the details of task planning and execution in Sec~\ref{sec:task_planning}.

\begin{figure*}[t]
    \centering
    \includegraphics[width=\linewidth]{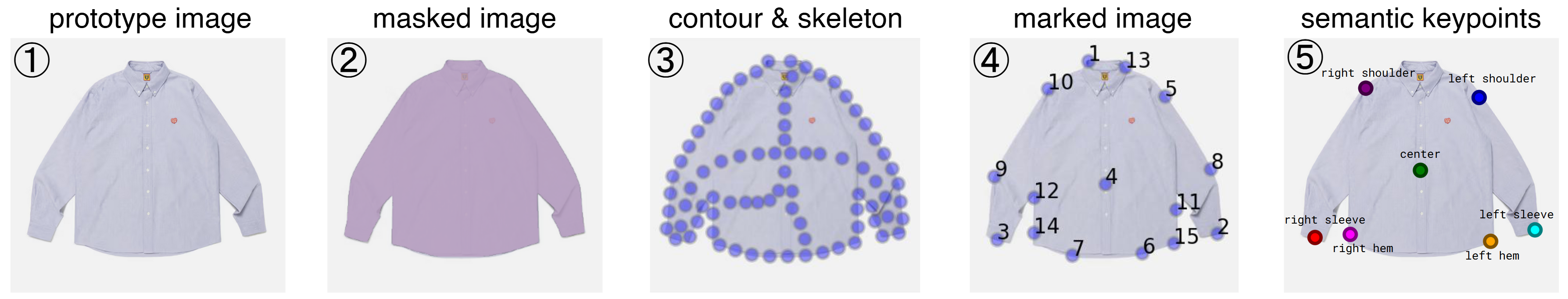}
    \caption{Semantic keypoint discovery. A fully-automated pipeline discovers semantic keypoints on a prototype image for each clothes type.}
     \label{fig:discovery}
    \vspace{-0.4cm}
\end{figure*}

\section{Semantic Keypoints}
\label{sec:keypoint_extraction}
The core idea of CLASP is to use semantic keypoints as a general state representation for clothes. Each semantic keypoint includes: (i) a language description (\eg, ``left sleeve"), providing semantic context for high-level task planning, and (ii) a keypoint position, offering geometric information to guide low-level action generation. Although previous work has explored using keypoints for clothes manipulation, most methods rely on manually annotated keypoints~\citep{fold_2, keypoint_synthetic}. Since clothes are highly deformable, collecting sufficient data that captures clothes' various configurations is time-consuming and labor-intensive. To mitigate data scarcity, augmenting 3D synthetic data in the simulation environment has been explored~\citep{clasp_icra}. However, this approach is difficult to scale to open-category clothes due to the limited availability of 3D clothes models. Recently, researchers have explored zero-shot keypoint extraction by leveraging powerful and generalizable feature maps from vision foundation models~\citep{rekep,dino_keypoint,imitation_keypoint}. However, these methods can not ensure that the resulting keypoints have explicit semantic meaning and can be described using natural language. In addition, using the vision foundation model for consistent keypoint extraction on clothes remains challenging, especially under self-occlusion and deformation.\par

\begin{figure*}[ht]
    \centering
    \vspace{0.3cm}
    \includegraphics[width=\linewidth]{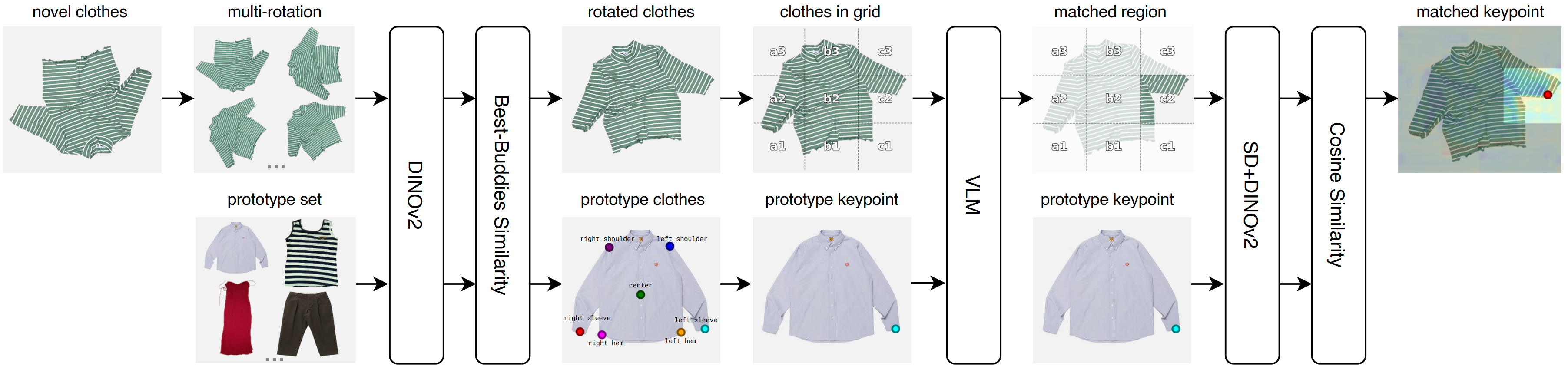}
    \caption{Semantic keypoint matching. Given an image of novel clothes, we first retrieve the most relevant prototype. Each semantic keypoint on the prototype is then matched to the novel clothes through a coarse-to-fine pipeline. Specifically, a VLM is employed for coarse region matching, while Stable Diffusion (SD) and DINOv2 are utilized for fine-grained keypoint matching.}
     \label{fig:keypoint}
\end{figure*}

Given that semantic keypoints encode both semantic and spatial information, two key technical challenges must be addressed in keypoint extraction: (1) understanding the semantics of open-category clothes and (2) maintaining the spatial precision of keypoints under deformation and occlusion. To address them, we propose a two-stage pipeline for keypoint extraction. The first stage (Figure~\ref{fig:discovery}) focuses on semantic understanding, aiming to autonomously discover semantic keypoints on prototype images and describe them using natural language. Specifically, we leverage geometric prior and the vision language model's commonsense knowledge to develop an autonomous semantic keypoint discovery pipeline. For each clothes category, we provide a prototype image free from occlusion and deformation, which makes semantic understanding feasible and also provides full observation as a reference in the second stage. The second stage (Figure~\ref{fig:keypoint}) focuses on spatial precision, aiming to match semantic keypoints on the prototype image to novel clothes, which may involve occlusions and deformations. Specifically, we utilize a coarse-to-fine matching strategy to propose semantic keypoints on novel clothes precisely.\par

\subsection{Keypoint Discovery}
\label{sec:keypoint_discovery}
For each clothes type, we provide a prototype image free from occlusion and deformation, allowing the semantic keypoint discovery stage to focus on understanding semantics and defining semantic keypoints for each clothes type. To discover semantic keypoints on the prototype image, we design an autonomous pipeline as shown in Figure~\ref {fig:discovery}, which leverages the geometric prior of clothes and the semantic understanding capability of the VLM for keypoint discovery. \par

To discover semantic keypoints, we first extract the contour and skeleton of clothes, which serve as an effective abstraction, capturing the geometric structure of clothes and encoding the human body. Specifically, we employ OWL-V2~\citep{OWLv2} to generate the bounding box of clothes on the prototype image based on the text prompt. Later, the bounding box will serve as the box prompts for the Segment-Anything Model (SAM)~\citep{SAM} to generate the corresponding masked image $I_m$ of the clothes. We then extract the contour of clothes from the masked image $I_m$ and perform skeletonization by calculating equidistance to the boundaries~\citep{skeleton_detect}, resulting in the skeleton.  To get keypoints from the skeleton and boundary, we then apply farthest point sampling (fps)~\citep{fps} on the boundary and skeleton to sample candidate keypoints $\Kc$. Farthest point sampling is an effective method for obtaining diverse, spatially distributed candidate keypoints that are geometrically salient. \par

After obtaining candidate keypoints $\Kc$, selecting semantic keypoints requires semantic understanding of clothes and clothes manipulation. Therefore, we leverage the commonsense knowledge of the VLM and employ mark-based visual prompting~\citep{moka} to select semantic keypoints and describe them with natural language. In this paper, we use GPT-4o as the VLM. The prompt is multi-modal and consists of two parts: (i) text prompt, and (ii) marked image. In the text prompt, we provide context information to the VLM, including the definition of semantic keypoints, the expected behavior of the VLM, and the desired response format. In the marked image, we draw visual marks on candidate keypoints $\Kc$ and label them with numbers. Through mark-based visual prompting, we can get the prototype semantic keypoints $\Kzero = \{\pzero,\, \szero\}_{i=1}^n$, where $\pzero$, $\szero$ denote the pixel position and language description of $i$-th semantic keypoint, respectively, and $n$ is the number of keypoints. \par

In this way, we establish a prototype set, where each image represents a clothes type and is annotated with semantic keypoints. 

\subsection{Keypoint Matching}
Although the autonomous pipeline can discover semantic keypoints on the prototype image in the first stage, it does not perform robustly on clothes with occlusion and deformation, which may lead to failures in boundary and skeleton extraction. Thus, the second stage performs semantic keypoint matching to transfer the semantic keypoints from the prototype image to novel clothes. The second stage ensures precise semantic keypoints extraction under occlusion and deformation.\par

The pipeline of semantic keypoint matching is shown in Figure~\ref{fig:keypoint}.
Given an RGB observation image of novel clothes, we first retrieve the most relevant prototype image by computing the Best-Buddies Similarity (BBS) using features extracted from the vision foundation model DINOv2~\citep{dinov2}. DINOv2 can extract rich and transferable visual features efficiently. BBS is a robust metric for measuring structural similarity between two images. It identifies pairs of local patches—one from the observation image and one from the prototype image—that are mutual nearest neighbors in the feature space. These pairs, referred to as best buddy pairs, capture strong local correspondences. The BBS score is defined as the proportion of local patches in the observation image that form such pairs. By evaluating the BBS score across all prototype images, we retrieve the one that is most structurally similar to the observation image. To enhance robustness, we apply a set of rotations to the observation image and select the rotation that yields the highest BBS score.\par
If the highest BBS score within the prototype image set falls below a predefined threshold, the observation is classified as a new clothes type, requiring a new prototype image. This new prototype image is then annotated using our autonomous semantic keypoint discovery pipeline. Otherwise, the prototype image with the highest BBS score is retrieved. \par

After retrieving the most relevant prototype image $I_0$ annotated with semantic keypoints, we then perform a coarse-to-fine semantic keypoint matching on novel clothes. Given the novel clothes' RGB observation image $I$ and a semantic keypoint position $\pzero$ on the prototype image, we first match $\pzero$ to a coarse region $\Omega$ on the observation image $I$. This stage is crucial for capturing global context and avoiding local minima. Specifically, we employ VLM with mark-based visual prompting for coarse region proposals. We first mark the observation image $I$ with an overlaid grid, where each grid cell is indexed by a unique text label, and mark the prototype image $I_0$ with a dot on $\pzero$. We then prompt the VLM with the marked observation and prototype image to select the grid indices corresponding to the prototype semantic keypoint position $\pzero$. To obtain precise semantic keypoint positions, we then perform fine-grained keypoint matching on the region $\Omega$ with vision foundation models. Borrowing the method from \citep{semantic_matching}, we fuse the feature maps from DINOv2~\citep{dinov2} and Stable Diffusion (SD)~\citep{stable-diffu}. The fused feature effectively captures both local and global information, thereby enhancing the robustness of semantic keypoint matching. Specifically, we feed the observation image $I$ and the prototype image $I_0$ into the fused feature extractor $\mathcal{F}(\cdot)$, 
which produces their corresponding fused feature maps $\mathcal{F}(I)$ and $\mathcal{F}(I_0)$. 
For each point $p \in \Omega$ on the observation image, we obtain its feature representation $\mathcal{F}(I,\, p)$, and for each prototype semantic keypoint $\pzero$, we obtain its corresponding prototype feature $\mathcal{F}(I_0,\, \pzero)$. We then compute the cosine similarity between $\mathcal{F}(I,\, p)$ and $\mathcal{F}(I_0,\, \pzero)$ to perform fine-grained keypoint matching:

\begin{equation}
    p_i = \arg\max_{p \in \Omega} 
    \langle \mathcal{F}(I,\, p),\, \mathcal{F}(I_0,\, \pzero) \rangle,
    \label{eq:matching}
\end{equation}

where $\langle \cdot, \cdot \rangle$ denotes the cosine similarity between two feature vectors. 
Here, $p_i$ represents the pixel position of the matched keypoint on the observation image $I$ corresponding to the prototype keypoint $\pzero$. Repeating this process for all prototype semantic keypoints $\Kzero$, we obtain the matched semantic keypoints $\K$ on the novel clothes. With the pixel locations of these keypoints, we further compute their 3D coordinates using the corresponding depth map and camera parameters.

\begin{figure*}[t]
    \centering
    \vspace{0.3cm}
    \includegraphics[width=\linewidth]{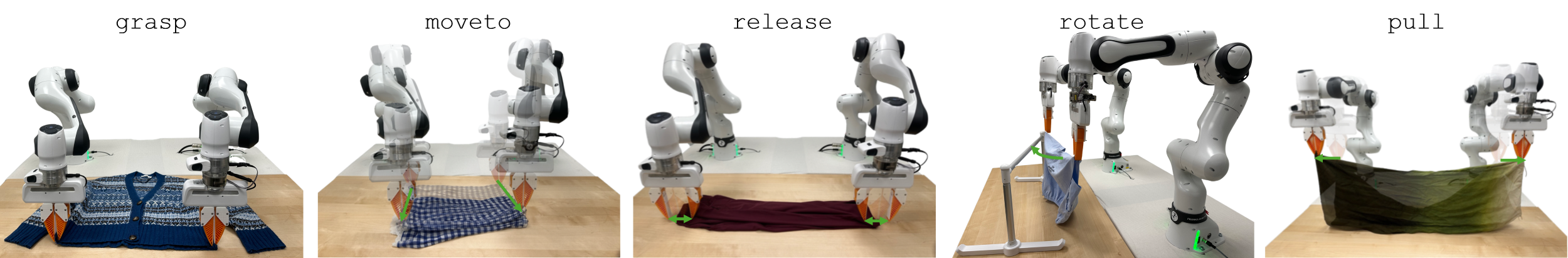}
    \caption{Skill library. CLASP skill library consists of 5 basic skills: grasp, moveto, release, rotate and pull. }
     \label{fig:action_prim}
\end{figure*}

\section{Clothes Manipulation With Semantic Keypoints}
\label{sec:clothes_manipulation}
\subsection{Skill Library}
\label{sec:basic_skill}

For general-purpose clothes manipulation, we build a skill library first. Instead of manually designing skills, we prompt an LLM to discover basic clothes manipulation skills. The LLM’s commonsense knowledge ensures these skills are general and sufficient for various clothes and manipulation tasks. Specifically, the LLM is prompted with a chain-of-thought strategy~\citep{cot} to (i) generate examples of clothes manipulation tasks, (ii) break down these tasks into basic actions, and (iii) summarize the actions identified in step (iii) to derive a set of basic skills. In this way, we identify the basic skill set $\mathcal{A}$, including \texttt{grasp}: pick up a part of the clothes, \texttt{moveto}: transport the clothes to a target position, \texttt{release}: drop a part of the clothes, \texttt{rotate}: rotate the clothes, and \texttt{pull}: stretch the clothes for flattening. Figure~\ref{fig:action_prim} visualizes the basic skills discovered by the LLM. Since semantic keypoints serve as the effective affordances for clothes manipulation and define where clothes are more likely to be manipulated, we implement these basic skills as policies parameterized by contact point positions. Specifically, we define heuristic rules to create waypoints based on the contact point positions for each basic skill. Once the waypoints are determined, a motion planning algorithm is applied to generate the complete trajectory for the robot's execution. Specifically, we implement each basic skill as follows:\par

    \noindent \texttt{grasp}$(\tl,\tr)$: The skill \texttt{grasp} picks up clothes parts using the grippers of one or both arms, where $\tl$ and $\tr$ denote the 3D coordinates of the grasping points for the left and right arms, respectively. Specifically, after getting 3D coordinates of grasping points from semantic keypoints, we will perform planar grasping on the table. The skill \texttt{grasp} supports single-arm and dual-arm modes, depending on the number of grasping points provided. The allocation of grasping points to the robot arms follows a greedy strategy, where each grasp point is assigned to the closer available arm.\par
    \noindent \texttt{moveto}$(\tl, \tr)$: The skill \texttt{moveto} transports the clothes to a target position, where $\tl$ and $\tr$ denote the 3D coordinates of the target positions for the left and right arms, respectively. Specifically, target positions are determined by semantic keypoints or the location of the receptacle (\eg, a rack or a box), depending on the task context. The location of the receptacle can be obtained through an open-vocabulary object detector. For the target orientation, the gripper is typically perpendicular to the tabletop in most cases; however, in hanging tasks, the gripper's orientation is adjusted to 30 degrees relative to the tabletop to expand the robot's workspace.\par
    \noindent \texttt{release}$()$: The skill \texttt{release} drops clothes parts by opening grippers. After the \texttt{release} action, the observation is updated, and CLASP extracts new semantic keypoints.\par
    \noindent
   \texttt{rotate}$(\theta)$: The skill \texttt{rotate} rotates the clothes grasped by the gripper, where $\theta$ represents the rotation angle within the table plane. The rotation axis is located at the center of the grippers of the two robot arms. This skill is frequently used to align the clothes with the rack in hanging tasks, and the rack's pose can be obtained using an open-vocabulary object detector and principal component analysis (PCA).\par
    \noindent
    \texttt{pull}$(l)$: The skill \texttt{pull} stretches the clothes to flatten them, where $l$ represents the stretch distance. In this paper, we set the stretch distance $l$ to 10\% of the distance between the grippers of the two arms. A potential improvement could involve stretching the clothes as much as possible without tearing, monitored through a force sensor or a front-view camera.

\subsection{Task Planning}
\label{sec:task_planning}

\begin{figure}[t]
    \centering
    \includegraphics[width=\linewidth]{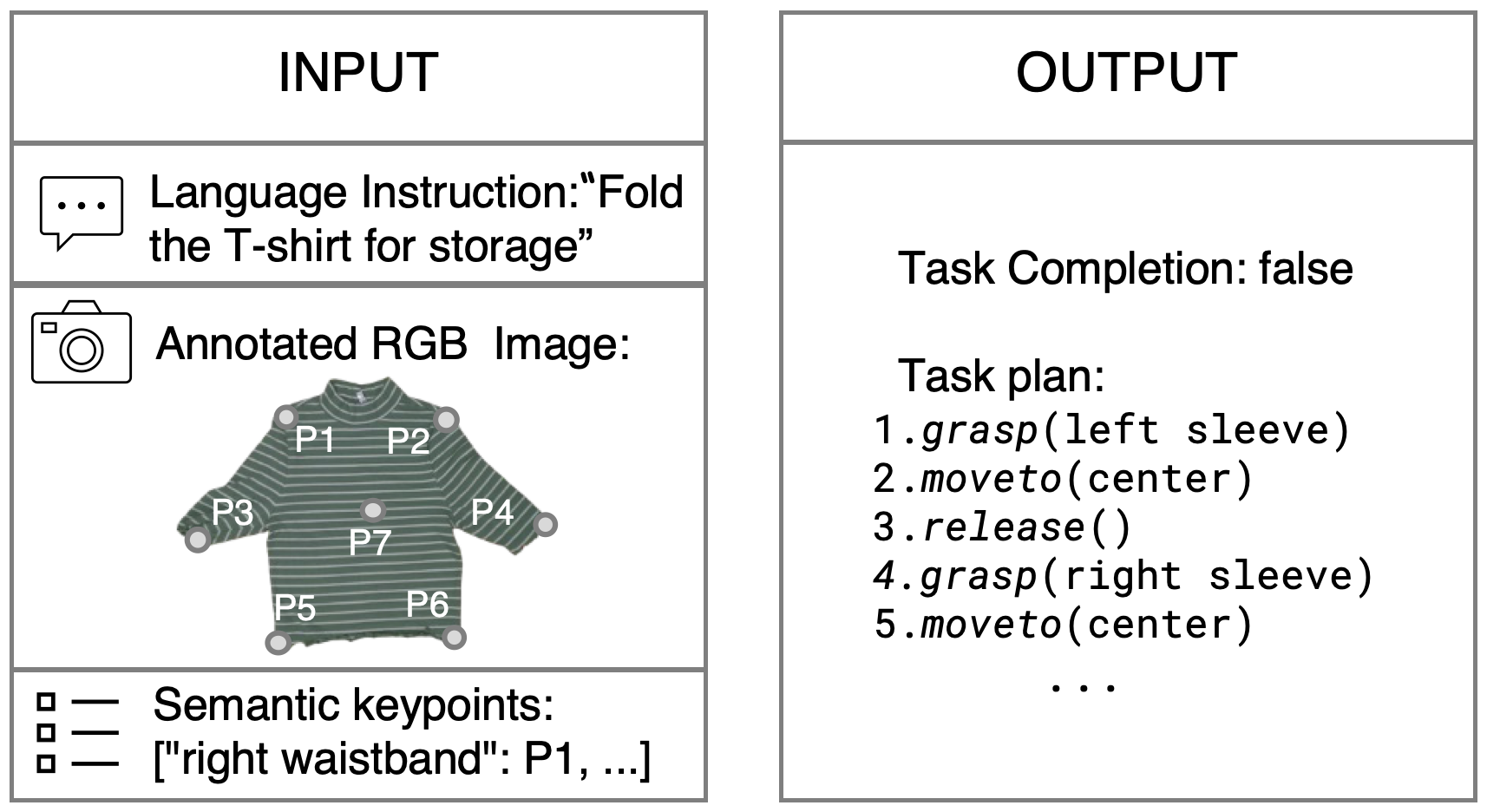}
    \caption{Example of few-shot prompting for the VLM-powered task planner.}
     \label{fig:task_plan}
    \vspace{-0.2cm}
\end{figure}

\begin{algorithm}[t]
\caption{Task Planning Procedure}
\label{alg:task_planning_overall}
\begin{algorithmic}[1]
    \Require Instruction $\La$, image $I$, keypoints $\K$, skill set $\A$
    \Ensure Valid task plan or failure reason

    \If{\textsc{CheckCompletion}($\La, I, \K$)}
        \State \Return \texttt{Task Completed}
    \EndIf

    \Repeat
        \State $\boldsymbol{W} \gets$ \textsc{ProposeTaskPlan}($\La, I, \K, \A$)
        \State $valid \gets$ \textsc{PlanVerification}($\boldsymbol{W}, \A, \K$)
    \Until{$valid$ or attempts are exhausted}

    \State \Return $\boldsymbol{W}$ or failure reason
\end{algorithmic}
\end{algorithm}

\begin{algorithm}[t]
\caption{Check validity of a proposed task plan $\boldsymbol{W}$}
\label{alg:plan_verification}
\begin{algorithmic}[1]
    \Statex \textsc{PlanVerification}($\boldsymbol{W}, \A, \K$)
    \For{each $\boldsymbol{w}_j = (a_j, c_j^1, c_j^2)$ in $\boldsymbol{W}$}
        \If{$a_j \notin \A$ or $c_j^{1,2} \notin \K$}
            \State \Return \texttt{False}
        \ElsIf{not \textsc{MotionFeasibility}($a_j, c_j^1, c_j^2$)}
            \State \Return \texttt{False}
        \EndIf
    \EndFor
    \State \Return \texttt{True}
\end{algorithmic}
\end{algorithm}

After establishing the skill library for clothes manipulation, we leverage the VLM to perform task planning over semantic keypoints, utilizing its commonsense knowledge acquired from internet-scale data~\citep{llm_planning}. The task planning procedure (Algorithm~\ref{alg:task_planning_overall}) consists of three main modules: \textsc{CheckCompletion}, \textsc{ProposeTaskPlan}, and \textsc{PlanVerification}.

\textsc{CheckCompletion}:
Given a language instruction $\La$, an observation image $I$, and extracted semantic keypoints $\K$, CLASP first determines whether the task specified by $\La$ has already been completed. If the completion condition is satisfied, the procedure terminates.

\textsc{ProposeTaskPlan}:
If the task remains incomplete, the VLM generates a candidate task plan $\boldsymbol{W} = (\boldsymbol{w}_1, \boldsymbol{w}_2, \ldots, \boldsymbol{w}_H)$. Each subtask $\boldsymbol{w}_j = (a_j, c_j^1, c_j^2)$ consists of a basic skill $a_j$ and up to two contact points $(c_j^1, c_j^2)$ for the dual-arm manipulation system. The skill $a_j$ is selected from the predefined set $\mathcal{A}$ introduced in Sec.~\ref{sec:basic_skill}, while the contact points are chosen from the semantic keypoints $\K$. Semantic keypoints define where clothes can be manipulated and serve as contact point candidates in high-level task planning. To improve robustness, we employ in-context learning by providing the VLM with a few-shot prompt that includes example pairs of natural-language instructions, observation images annotated with semantic keypoints, and corresponding task plans (Figure~\ref{fig:task_plan}). These examples define the expected input–output structure and demonstrate how to ground instructions into executable skills over semantic keypoints.

\textsc{PlanVerification}:
The proposed candidate task plan $\boldsymbol{W}$ is then validated through the procedure described in Algorithm~\ref{alg:plan_verification}. For each subtask $\boldsymbol{w}_j$, CLASP first performs syntax and semantic checks—verifying that the skill $a_j$ exists in $\mathcal{A}$ and that contact points $c_j^{1,2}$ correspond to valid semantic keypoints. Next, it evaluates geometric feasibility via motion planning. A subtask passes verification if a feasible trajectory is found; otherwise, the failure reason is used to prompt VLM to generate a new task plan.\par

Once a verified task plan is obtained, CLASP executes the subtasks sequentially. After each subtask, the RGB-D observation is updated. However, due to potential occlusion caused by the robot, semantic keypoints are re-extracted only after executing a \texttt{release} action. Following each keypoint update, the VLM is prompted to replan based on the updated observation and keypoints. This process repeats iteratively until the task is successfully completed.

\section{Experiments}
\label{sec:experiments}

\begin{figure*}[ht]
    \centering
    \includegraphics[width=\linewidth]{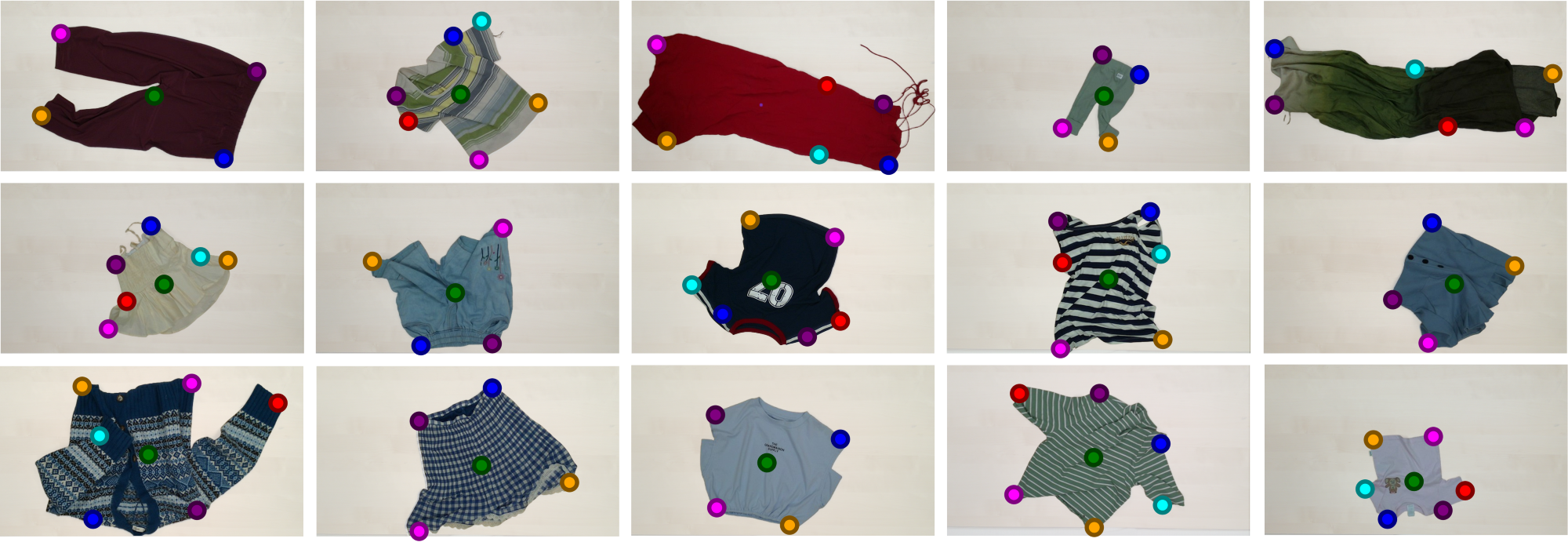}
    \caption{Qualitative results on semantic keypoint extraction. Our method can achieve open-category semantic keypoint extraction under irregular deformation and occlusion.}
    \label{fig:kp_resu}
\end{figure*}

Our experiments investigate three  questions: \par
Q1. Can CLASP’s submodules perform effectively under varying conditions? \par
We evaluate the performance of CLASP’s submodules: semantic keypoint extraction and task planning. Sec.~\ref{subsec:keypoint_expri} shows that the proposed method reliably extracts semantic keypoints across various clothes, even under occlusion and deformation. Sec.~\ref{subsec:task_plan} shows that the task planner generates robust, executable plans, with closed-loop planning substantially improving performance on crumpled clothes.\par
Q2. How well does CLASP perform on clothes manipulation compared with the existing baseline method?  \par
We address this question through comparative experiments detailed in Sec.~\ref{subsec:simu_expri}. CLASP outperforms baseline methods in terms of manipulation success rate and generalization capability, highlighting its effectiveness.\par

Q3. Can CLASP generalize to a wide range of clothes types and manipulation tasks in the real world? \par
To answer this, we conduct extensive real-world evaluations, as presented in Sec.~\ref{subsec:real_expri}. Results show that CLASP generalizes well to various clothes types and manipulation tasks, confirming its practical applicability.\par

\subsection{Semantic Keypoint Extraction Experiments}
\label{subsec:keypoint_expri}

\begin{table}[ht]
\centering
\caption{Semantic keypoint extraction performance.}
\resizebox{0.95\linewidth}{!}{
\begin{tabular}[t]{ccccc}
\toprule
Expert &AKD (pixel) &$\textit{AP}_{15}$ (\%) 
&$\textit{AP}_{30}$ (\%)  &$\textit{AP}_{45}$ (\%) 
\\
\midrule
1 &16.2 &71.0&88.0&90.7  \\
2 &17.8 &69.0&84.1&90.4\\
3 &19.7 &68.7&82.2&87.8\\
\bottomrule
\end{tabular}}
\label{tab:keypoint}
\end{table}%

We first evaluated the performance of CLASP in extracting effective semantic keypoints. Since no ground-truth semantic keypoints are available, we provide annotated prototype images to three human experts and ask them to label semantic keypoints on images of novel clothes. In this way, we establish a testing set of 120 images from 15 different clothes. In these images, clothes have diverse configurations, poses, occlusions, and deformations. We then compare the semantic keypoints extracted by CLASP against those labeled by the experts. \par

The evaluation utilizes two standard metrics for keypoint extraction: Average Keypoint Distance (AKD) and Average Precision (AP). AKD calculates the average distance between the extracted keypoints and the ground-truth keypoints, while AP measures the proportion of correctly extracted keypoints under different thresholds. Since the resolution of the observed RGB-D images is 720×1280, we used thresholds of 15, 30, and 45 pixels for AP calculation.\par

The results are shown in TABLE~\ref{tab:keypoint}. The average keypoint distance between the semantic keypoints proposed by human experts and our method is less than 20 pixels, while the average precision under the 30 pixels threshold is more than 80\%. 

Considering the resolution and uncertainty in human expert labeling, such as identifying the center of clothes, the results can demonstrate that our proposed method performs well in extracting semantic keypoints. 
 Figure~\ref{fig:kp_resu} shows some qualitative results of semantic keypoint extraction. The experiment results demonstrate that our two-stage semantic keypoint extraction pipeline is effective in extracting semantic keypoints of clothes even under occlusion and deformation. The prior knowledge from the VLM and vision foundation models ensures the generalization to diverse clothes types. Besides, coarse-to-fine keypoint matching ensures spatial precision and makes semantic keypoint extraction robust. Our method enables effective semantic keypoint extraction on open-category clothes using a single unannotated prototype image per clothes category, without the need for large-scale labeled data.

\begin{table}[t]
\centering
\caption{Task planning performance.}
\resizebox{\columnwidth}{!}{
\begin{tabular}{lcccccc}
\toprule
\multirow{2}{*}{Task} 
& \multicolumn{2}{c}{Check Completion (\%)} 
& \multicolumn{3}{c}{Subtask Accuracy (\%)} \\
\cmidrule(lr){2-3} \cmidrule(lr){4-6}
& VLM & VLM + KP & LLM & VLM & VLM + KP  \\
\midrule
Folding      & 66.7  & 93.3 & 33.3 & 56.7  & 86.7   \\
Flattening   & 73.3  & 96.7 & 50.0 & 60.0  & 93.3   \\
Hanging      & 93.3  & 100.0 & 96.7 & 76.7  & 100.0  \\
Placing      & 83.3  & 96.7 & 96.7 & 70.0  & 96.7   \\
\bottomrule
\end{tabular}
}
\label{tab:task_planning_results}
\end{table}
\vspace{-0.15cm}

\subsection{Task Planning Experiments}
\label{subsec:task_plan}
To evaluate the performance of the proposed task planner for clothes manipulation, we collect images annotated with ground-truth semantic keypoints (labeled by human experts) along with free-form natural language descriptions to evaluate the planner on folding, flattening, hanging, and placing tasks. For each task category, we collect 30 images of 15 diverse clothes. We evaluate the planner using two metrics: check completion accuracy, which assesses whether the planner correctly determines if the overall task has been completed based on the observation; and subtask accuracy, which measures whether the generated subtask sequence is appropriate for the observed state. We compare our method (VLM + KP) with two baselines: (i) a VLM baseline that takes raw images without semantic keypoints, and (ii) an LLM baseline that receives only the language descriptors of semantic keypoints along with the language instruction, without any visual input (as in our previous work~\cite{clasp_icra}). All methods receive the same few-shot demonstrations but differ in visual input: our method uses keypoint-annotated images, the VLM baseline uses raw images, and the LLM baseline does not rely on any visual input.\par

Table~\ref{tab:task_planning_results} presents the experimental results. Since the LLM baseline lacks visual input, we do not report its accuracy in checking completion. It performs the worst on folding and flattening tasks, where adaptation to varied states is crucial. For folding, some parts of the clothes (such as sleeves) may be folded or unfolded after execution. For flattening, the model needs to understand the state of the clothes from visual observation and determine whether flattening is required. For example, in some flattening tasks, the robot must repeatedly flatten the clothes to achieve a flat state. The LLM baseline, being an open-loop planner, generates task plans solely from symbolic descriptions without observing the outcomes of action execution. This open-loop nature leads to failures in tasks that require dynamic replanning. In contrast, our planner operates in a closed-loop manner, continuously updating the plan based on visual and keypoint feedback to adjust its strategy when action outcomes deviate from expectations. The LLM baseline performs slightly better in hanging and placing tasks, which involve less uncertainty in the initial state. Moreover, the LLM’s symbolic reasoning remains consistent and unaffected by visual distribution shifts.\par

Compared to the VLM baseline without semantic keypoints, our pipeline achieves more reliable and interpretable task planning, especially in long-horizon folding tasks. Without semantic keypoints, the VLM often hallucinates or generates incorrect plans. Semantic keypoints provide critical structural information about the clothes’ state, allowing the model to reason over geometry and topology rather than surface appearance. This enhances robustness under distribution shifts between the observed images and the few-shot examples in the prompt.\par

Overall, our proposed planner achieves reliable accuracy in generating appropriate task plans and determining task completion. By providing semantic keypoints and a predefined skill library, we enable structured and controllable planning. To further enhance robustness, we employ in-context learning with examples that include language instructions, observation images with semantic keypoint annotations, and corresponding task plans. All generated plans are also verified before execution, and the planner supports dynamic replanning based on new observations, resulting in stable and robust closed-loop task planning.\par

\begin{table*}[t]
\vspace{0.2cm}
\centering
\small
\caption{Simulation experiments. The average success rates (\%) on testing tasks. The best performance is in bold.}
\resizebox{\linewidth}{!}{
\begin{tabular}{@{}lc cc cc cc cc @{}}
\toprule
\multirow{2}*{Method} 
& \multicolumn{2}{c}{\begin{tabular}[c]{@{}c@{}} Folding
\\(seen object) \end{tabular}} 
& \multicolumn{2}{c}{\begin{tabular}[c]{@{}c@{}} Flattening
\\ (seen object) \end{tabular}} 
& \multicolumn{2}{c}{\begin{tabular}[c]{@{}c@{}} Hanging
\\ (seen object) \end{tabular}} 
& \multicolumn{2}{c}{\begin{tabular}[c]{@{}c@{}} Placing
\\ (seen object) \end{tabular}}
\\
\cmidrule(lr){2-3} \cmidrule(lr){4-5} \cmidrule(lr){6-7} 
\cmidrule(lr){8-9} 
& Towel &T-shirt &T-shirt &Skirt &Trousers &Towel &Towel &Skirt \\
\midrule
CLIPORT~\citep{languge_conditioned_2} &77.5 &80.0 &32.5 &36.7 &76.7 &83.3 &93.3 &93.3 \\
Goal-conditioned Transporter~\citep{goal_conditioned_2} &83.3 &76.7 &26.7 &33.3 &\textbf{100.0} &80.0 &66.7 &83.3 \\
FlingBot~\citep{flingbot} &N/A &N/A &\textbf{66.7} &\textbf{85.0} &N/A &N/A &N/A &N/A \\
FabricFlowNet~\citep{fabricflownet} &93.7 &\textbf{100.0} &N/A &N/A &N/A &N/A &N/A &N/A \\
CLASP  & \textbf{100.0} & 95.0 & 65.0 & 80.0 & 96.7 & \textbf{97.5} & \textbf{96.7} & \textbf{96.7} \\
\midrule
\multirow{2}*{Method} 
& \multicolumn{2}{c}{\begin{tabular}[c]{@{}c@{}} Folding
\\(unseen object) \end{tabular}} 
& \multicolumn{2}{c}{\begin{tabular}[c]{@{}c@{}} Flattening
\\ (unseen object) \end{tabular}} 
& \multicolumn{2}{c}{\begin{tabular}[c]{@{}c@{}} Hanging
\\ (unseen object) \end{tabular}} 
& \multicolumn{2}{c}{\begin{tabular}[c]{@{}c@{}} Placing
\\ (unseen object) \end{tabular}}  \\
\cmidrule(lr){2-3} \cmidrule(lr){4-5} \cmidrule(lr){6-7} 
\cmidrule(lr){8-9} 
&Trousers &Skirt &Trousers &Towel &T-shirt &Skirt &Trousers &T-shirt\\
\midrule
CLIPORT~\citep{languge_conditioned_2} &0.0 &0.0 &8.3 &9.2 &76.7 &66.7 &70.0 &73.3 \\
Goal-conditioned Transporter~\citep{goal_conditioned_2} &0.0 &0.0 &10.0 &6.7 &36.7 &40.0 &36.7 &60.0 \\
FlingBot~\citep{flingbot} &N/A &N/A & 29.2 & 34.2 &N/A &N/A &N/A &N/A \\
FabricFlowNet~\citep{fabricflownet} &0.0 &2.5 &N/A &N/A &N/A &N/A &N/A &N/A \\
CLASP &\textbf{87.5} &\textbf{81.7} &\textbf{60.8} &\textbf{65.0} &\textbf{93.3} &\textbf{76.7} &\textbf{93.3} &\textbf{94.2} \\

\bottomrule
\end{tabular}}
\label{tab:exp_resu}
\vspace{-0.25cm}
\end{table*}

\subsection{Simulation Experiments}
\label{subsec:simu_expri}

To evaluate CLASP's performance on clothes manipulation, we conduct experiments in SoftGym, where clothes are modeled as particles with ground-truth positions. We sample 3D meshes of clothes from CLOTH3D~\citep{cloth3d} dataset, covering four common clothes types in human's daily life: T-shirts, trousers, skirts, and towel. For each type, over 35 instances of varying sizes and shapes are used. In addition, we extend the SoftGym benchmark to 16 clothes tasks. These tasks can be divided into 4 types:\par

\begin{itemize}
    \item \textbf{Folding} tasks involve folding clothes to achieve a goal configuration. The success of a folding task is determined by the mean particle position error between the achieved and goal configuration.\par
    \item  \textbf{Flattening} tasks involve flattening crumpled clothes with random deformations and occlusion. The success of a flattening task is determined by the coverage area of the clothes. \par
    \item \textbf{Hanging} tasks require hanging clothes on a rack. A hanging task is successful when the clothes are fully hung on the rack without any part touching the ground. \par
    \item \textbf{Placing} tasks require placing the clothes in a box.  A placing task is successful when the clothes are laid flat inside the box.\par  
\end{itemize}

Baseline methods include two general multi-task learning frameworks (CLIPORT and Goal-conditioned Transporter) and two task-specific algorithms (FlingBot and FabricFlowNet): \par

\begin{itemize}
    \item \textbf{CLIPORT}~\citep{languge_conditioned_2} represents a typical end-to-end algorithm for learning language-conditioned manipulation policies, which leverages a pretrained vision language model and a two-stream architecture for semantic understanding and spatial precision.
    \item \textbf{Goal-conditioned Transporter}~\citep{goal_conditioned_2} represents a typical goal-conditioned transporter network for deformable object manipulation, which infers the optimal action from the current and goal images with a TransporterNets structure.
    \item \textbf{FlingBot}~\citep{flingbot} is a self-supervised learning framework designed for flattening crumpled clothes by using an effective fling action.
    \item \textbf{FabricFlowNet}~\citep{fabricflownet} is a learning framework designed for clothes folding, which infers the optimal action based on optical flow.
\end{itemize}

For each baseline method, we provide 1000 expert demonstrations per task for model training. To evaluate the generalization of baseline methods, only half of the tasks are seen during training through expert demonstrations. For CLASP, we provide few-shot examples of seen tasks in the prompt. Unseen tasks involve new object types. For each task, we conduct 120 trials with different clothes configurations to calculate the success rate. Two task-specific baseline methods are designed for clothes folding and flattening, respectively. We only test them on related tasks, and other tasks are marked as N/A (Not Applicable).\par

The experiment results are shown in TABLE~\ref{tab:exp_resu}. Overall, CLASP outperforms the two multi-task learning methods on both seen and unseen tasks. Compared with Goal-conditioned Transporter, CLIPORT shows better generalization. Language instructions facilitate capturing similarities between different tasks at the symbolic level. Besides, the pretrained vision-language model enables CLIPORT to capture such similarities. However, CLIPORT's generalization is limited when adapting to long-horizon tasks, such as transferring the skill from folding a T-shirt to folding trousers, because it learns task-specific manipulation action sequences in an end-to-end manner, and the learned action sequences are not transferable. In contrast, CLASP learns task-agnostic and generalizable language and visual concepts across a wide range of clothes manipulation tasks. The commonsense knowledge from VLM allows CLASP to handle clothes manipulation tasks by decomposing them into predefined basic skills. Furthermore, semantic keypoints 
are a general spatial-semantic representation of clothes, which are task-agnostic and provide effective cues for task planning and action generation in clothes manipulation.\par
Compared to the two task-specific algorithms, CLASP shows comparable performance on seen clothes folding and flattening tasks, demonstrating the effectiveness of the proposed method for clothes manipulation. However, on unseen clothes types, two task-specific algorithms tend to fail due to distribution shifts in observations. In contrast, CLASP is a general-purpose clothes manipulation method.\par

\begin{figure*}[t]
    \centering
    \includegraphics[width=\linewidth]{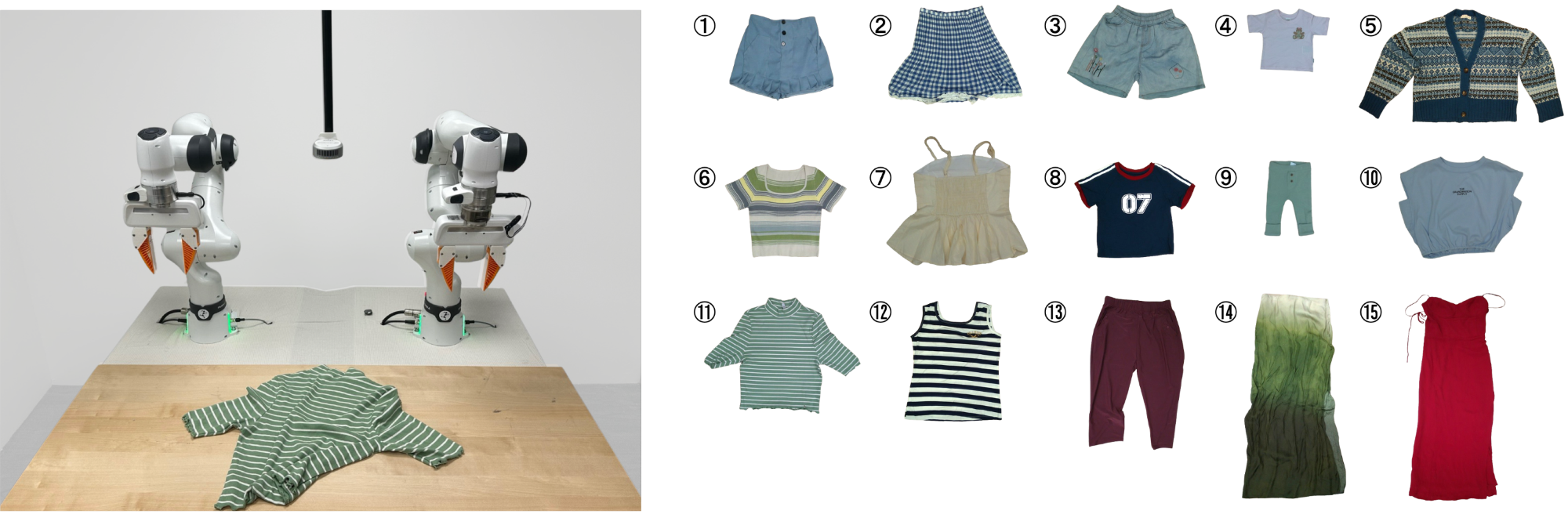}
    \caption{Robot experiments. The experimental setup consists of a Franka dual-arm system performing folding, flattening, hanging, and placing over many different clothes types.}
    \label{fig:system}
\end{figure*}

\begin{table*}[ht]
\vspace{0.2cm}
\centering
\small
\caption{Robot experiments. The success rates on various clothes and tasks. }
\resizebox{\linewidth}{!}{
\begin{tabular}{lcccccccccccccccc}
\toprule
\multirow{2}{*}{Task} & \multicolumn{16}{c}{{ Clothes ID}} \\
 \cmidrule(lr){2-17}
 & 1 & 2 & 3 & 4 & 5 & 6 & 7 & 8 & 9 & 10 & 11 & 12 & 13 & 14 & 15 & \textbf{Mean} \\
 \midrule
Folding & 4/5 & 5/5 & 4/5 & 4/5 & 5/5 & 5/5 & 4/5 & 5/5 & 3/5 & 4/5 & 4/5 & 5/5 & 5/5 &3/5 & 4/5 & 4.3/5 \\
Flattening & 2/5 & 4/5 & 4/5 & 3/5 & 3/5 & 4/5 & 2/5 & 5/5 & 2/5 & 3/5 & 3/5 & 4/5 & 3/5 & 3/5 & 4/5 & 3.3/5 \\
Hanging & 5/5 & 5/5 &5/5 & 5/5 & N/A & 5/5 & 4/5 & 5/5 & 5/5 & 4/5 & 4/5 & 5/5 & N/A & N/A & N/A & 4.7/5\\
Placing & 4/5 & 5/5 & 5/5 & 5/5 & N/A & 5/5 & 5/5 & 5/5 & 5/5 & 4/5 & 3/5 & 5/5 & N/A & N/A & N/A & 4.6/5 \\
\bottomrule
\end{tabular}}
\label{tab:exp_real}
\end{table*}

\subsection{Real-world Experiments}
\label{subsec:real_expri}

To evaluate the performance and generalization of CLASP in real-world scenarios, we establish a dual-arm robot manipulation system. As shown in Figure~\ref{fig:system}, the system consists of two Franka Research 3 robot arms and a top-down RealSense L515 RGB-D camera for capturing RGB-D images. The clothes are placed on a table in front of the robot. We use SAM~\citep{SAM} and OWLv2~\citep{OWLv2} to segment the clothes for semantic keypoint extraction. To generate the dual-arm trajectories, we implement a motion planning algorithm using MoveIt!~\citep{moveit} to avoid collisions and synchronize the motion between the arms.\par

In real-world experiments, we evaluate CLASP on 15 clothes across diverse types, such as shirts, T-shirts, sweaters, skirts, long dresses, and vests, size (ranging from 18 cm × 30 cm for baby pants to 42 cm × 101 cm for adult long dresses), shapes (long sleeves, short sleeves, and sleeveless), and materials (\eg, cotton, polyester, and wool). We directly evaluate CLASP's performance on manipulation tasks of these clothes. Each clothes item is evaluated across four tasks: folding, flattening, hanging, and placing. The evaluation process includes five trials per task, with success rates calculated as follows. Folding: Success is determined by the Intersection over Union (IoU) between cloth masks achieved by CLASP and human experts. If the IoU is 90\% or higher, the task is considered successful. Flattening: if the flattened clothes cover at least 90\% of their maximum possible area in a fully flat state, the flattening task is successful. Hanging: the task is considered successful if the clothes are completely hung on the rack without any part touching the ground. Placing: the task is considered successful when the clothes are entirely placed within the box.\par

\begin{figure*}[ht]
    \centering
    \vspace{1cm}
    \includegraphics[width=\linewidth]{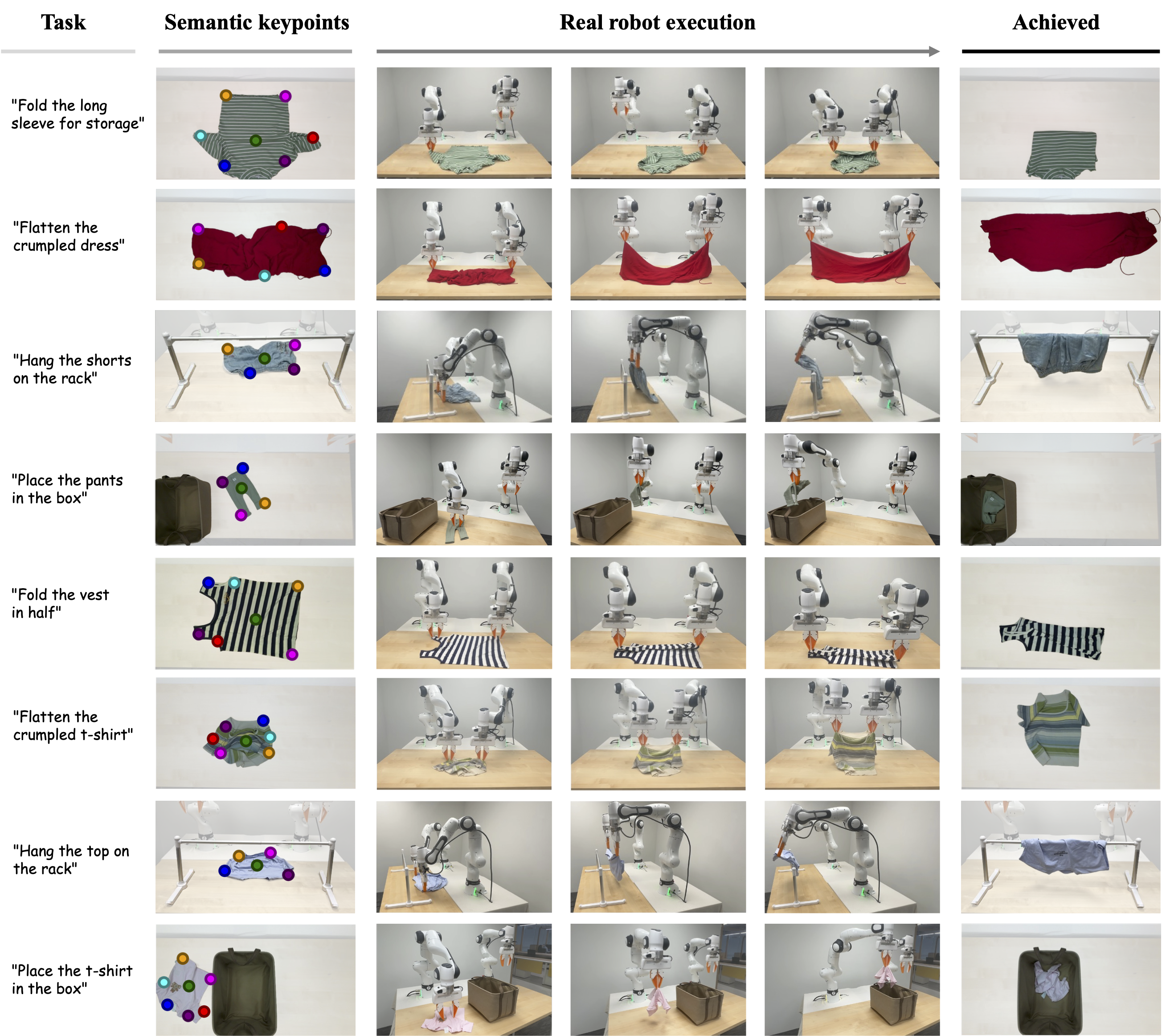}
    \caption{Qualitative results of real-world experiments. From left to right, the figure displays the task description, extracted semantic keypoints, the real robot's execution process, and the final achieved state.}
    \label{fig:experiment}
\end{figure*}

The experiment results are shown in Table~\ref{tab:exp_real}. For clothes 5, 13, 14, and 15, due to limitations in the robot's working space, we did not perform evaluations on hanging and placing tasks. These are marked as N/A (Not Applicable). Overall, CLASP can be applied to a wide variety of clothes and manipulation tasks while achieving a high success rate. CLASP achieves an 86\% success rate in clothes folding, a 66\% success rate in clothes flattening, a 94\% success rate in clothes hanging, and a 92\% success rate in clothes placing. The success rate is comparable to existing task-specific clothes manipulation algorithms, while we test CLASP on broader clothes types and instances. With a general semantic keypoints representation and the prior knowledge from foundation models, CLASP is an effective general-purpose clothes manipulation method. Fig.~\ref{fig:experiment} illustrates some representative examples. For each task, we visualize the extracted semantic keypoints, manipulation action sequences, and achieved object state. CLASP demonstrates the ability to manipulate various clothes in different ways in real-world scenarios. More videos of real-world experiments can be found in our supplementary material.\par

\section{Discussion}
\label{sec:discussion}
Despite the promising overall results, CLASP still shows several limitations based on the experimental results. First, CLASP struggles to extract accurate semantic keypoints of clothes with significant occlusion or deformation. This limitation arises because such cases are underrepresented in the datasets of the vision foundation models and VLMs that CLASP relies on. Although CLASP can perform flattening based on semantic keypoints to reduce occlusion, its success rate is lower than in other tasks (Table~\ref{tab:exp_resu} and Table~\ref{tab:exp_real}). Nevertheless, it achieves such performance and generalization without using any robot manipulation data. Future work could improve robustness by refining the policy through imitation learning with additional demonstrations. Second, CLASP faces challenges in manipulating large clothes due to workspace limitations. In real-world experiments, it sometimes fails to generate geometric feasible task plans for large clothes, preventing evaluation on certain hanging and placing tasks (\eg, clothes 5, 13, 14, and 15).  This limitation could be addressed by integrating a mobile manipulator to expand the reachable workspace~\citep{equivact} or by adding dragging skills~\citep{Toussaint}, allowing CLASP to handle larger clothes with random pose. Lastly, during manipulation, sliding or deformation can affect performance, especially for silk clothes (clothes 14 in real-world experiments). The folding success rate remains relatively low due to excessive sliding and deformation. This issue could be addressed by closed-loop action execution based on real-time semantic keypoint tracking. However, since CLASP’s keypoint extraction depends on large foundation models, real-time perception is still challenging. A practical improvement would be to use CLASP’s current pipeline to generate labeled data and distill a smaller, lightweight keypoint extractor, enabling real-time keypoint tracking and online trajectory optimization to enhance action robustness.
\section{Conclusion}
\label{sec:conclusion}
In this paper, we present semantic keypoints as a general spatial-semantic representation of clothes. Building on this representation, we propose CLothes mAnipulation with Semantic keyPoints (CLASP), where semantic keypoints play a crucial role in both high-level task planning and low-level action generation. By integrating the general semantic keypoint representation with foundation models pretrained on internet-scale data, CLASP serves as a general-purpose method for clothes manipulation. Simulation experiments demonstrate that our method outperforms baseline methods in terms of success rate and generalization in clothes manipulation. Real-world experiments further validate CLASP's generalization in clothes manipulation. CLASP can be directly applied to a diverse range of clothes types and manipulation tasks in real-world scenarios. 


\onecolumn
\appendix
\section{ Discussion of Differences with the Conference Version}
\label{sec:app_difference}

 This manuscript is an extension of our previous work presented at the IEEE International Conference on Robotics and Automation (ICRA), titled ``General-purpose Clothes Manipulation with Semantic Keypoints''~\citep{clasp_icra}. We have significantly expanded upon the conference version in three primary aspects:
 \begin{itemize}
     \item[1.] Introduction of a new semantic keypoint extraction pipeline for open-category clothes, enhancing applicability and generalization (Section~\ref{sec:keypoint_extraction}).
    \item[2.] Introduction of a closed-loop task planning pipeline capable of plan verification, dynamic replanning based on new observations, and task completion checking, enhancing the ability to handle unexpected state changes of clothes (Section~\ref{sec:task_planning}).
     \item[3.] Expansion of real-world experimental evaluations to diverse clothes categories, demonstrating the generality and robustness of our framework (Sec~\ref{subsec:real_expri}).
 \end{itemize}

 \subsection{Semantic Keypoint Extraction for Open-category Clothes}
In the conference version, we trained a model to extract semantic keypoints from observations. However, due to the high-dimensional configuration space of clothes, collecting sufficient data is challenging. To mitigate data scarcity, we imported 3D clothes meshes into the simulation for data augmentation. Nevertheless, the limited availability of 3D mesh models restricted the model to extracting semantic keypoints from only four types of clothes. In this manuscript, we propose a novel semantic keypoint extraction pipeline for open-category clothes~(Sec.~\ref{sec:keypoint_extraction}). The proposed pipeline leverages large vision-language models and vision foundation models to overcome data scarcity and eliminate the need for training on labeled data. In this new pipeline, introducing a new clothes category requires only a prototype image—without any annotations—for autonomous semantic keypoint discovery (Sec.~\ref{sec:keypoint_discovery}). This enhancement significantly improves the applicability and generalization of the proposed method.

 \subsection{Closed-loop Task Planning}
In the conference version, task planning was performed in symbolic space—without conditioning on visual observations—and executed only once at the beginning of the manipulation task. This made the system sensitive to unexpected state changes, which are common when the initial clothes state is highly crumpled, as in flattening tasks. In this manuscript, we propose a closed-loop task planning pipeline (Sec.~\ref{sec:task_planning}), where the task planning is conditioned on visual observations. After each sub-task is executed, the observation is updated, and the pipeline determines whether to replan. Additionally, each generated plan is verified before execution, and the system continuously checks whether the overall task has been completed. If the task is not completed, a new plan is generated based on the current observation. This closed-loop task planning pipeline enhances the system’s ability to handle unexpected state changes of clothes. Experimental results in Sec.~\ref{subsec:task_plan} show that this enhancement significantly improves the system's performance when the initial clothes state is crumpled.

 \subsection{Expanded Real-world Experimental Evaluation}
Beyond the qualitative real-world results on four clothing categories presented in the conference version, we have extended our real-world experiments to include quantitative evaluations and analysis across a broader range of clothes categories, varying in size, shape, and material (Sec.~\ref{subsec:real_expri}). These new results further demonstrate the effectiveness of our proposed clothes manipulation method, which demands a higher level of generalization and robustness.

By introducing a novel open-category semantic keypoint extraction pipeline and a closed-loop task planning system, this manuscript presents substantial extensions over the conference paper.

\section{Prompts}

\begin{center}
\begin{tcolorbox}[colback=gray!5, colframe=black!40, sharp corners=south, title= Basic Skill Discovery]\small

Define basic skills for a dual-arm home-service robot, which is required to perform clothes manipulation tasks like folding, flattening, hanging, and placing.\\

Think about this step by step:
\begin{enumerate}[leftmargin=*, label=\arabic*.]
    \item provides some examples of clothes manipulation tasks.
    \item breaks down these tasks into basic actions.
    \item summarizes basic actions and derive a set of basic skills.
\end{enumerate}
\end{tcolorbox}
\end{center}

\newpage

\begin{center}
\begin{tcolorbox}[colback=gray!5, colframe=black!40, sharp corners=south, title= Semantic Keypoint Discovery]\small

Select the 'semantic keypoints' of clothes from 'candidate keypoints' to solve the clothes manipulation tasks. Don't ignore any semantic keypoints. \\

The input request is an image of the clothes, annotated with a set of visual marks: 
\begin{itemize}
    \item \textbf{candidate keypoints}:  black dots marked as 'P[i]' on the image, where [i] is an integer.\\
\end{itemize}

The definition of \textbf{semantic keypoints}: keypoints that are crucial for clothes manipulation tasks like folding, flattening, and hanging, these keypoints can be described with natural language. Please note: \par
\begin{itemize}
    \item In the folding process, both picking and placing keypoints are 'semantic keypoints'. Don't ignore any semantic keypoints.
    \item Don't ignore the center if the center is the placing point.
    \item clothes often have symmetric structures, which can be useful guidance.\\
\end{itemize}
 
The response should be a list of 'semantic keypoints', each semantic keypoint should be a dictionary that contains:
\begin{itemize}
    \item \textbf{keypoint}: P1
    \item  \textbf{language description}: `left shoulder'
    \item  \textbf{reason}: `crucial for symmetry and folding the top part along with the left shoulder'
\end{itemize}

\end{tcolorbox}
\end{center}

\begin{center}
\begin{tcolorbox}[colback=gray!5, colframe=black!40, sharp corners=south, title= Coarse Region Matching]\small

Match the region in the test image that corresponds to the specified keypoint in the template image.\\

The input request contains:  
\begin{itemize}
    \item \textbf{template image} annotated with a keypoint. The keypoint is marked as a red dot. 
    \item \textbf{test image} that is divided into several regions annotated with a set of visual marks. \\
\end{itemize}

The response should be the mark of the corresponding region in the following format:\\
The region corresponding to the red keypoint is ** **.

\end{tcolorbox}
\end{center}

\begin{center}
\begin{tcolorbox}[colback=gray!5, colframe=black!40, sharp corners=south, title= Sub-task Proposal]\small

Describe the dual-arm robot's actions to solve the clothes manipulation tasks by:
\begin{itemize}
\item selecting contact points from annotated semantic keypoints.
\item selecting basic skills from a pre-defined skill library.\\
\end{itemize}

You have the following basic skills to select: 
\begin{itemize}
    \item grasp(): pick up clothes parts using the grippers of one or both arms, the parameter can be one or two contact points.
    \item moveto(): transports the clothes or clothes parts to a target position, the parameter can be one or two contact points.
    \item release(): drops clothes parts by opening grippers.
    \item rotate(): rotates the clothes grasped by the gripper.
    \item pull(): stretches the clothes to flatten them.\\
\end{itemize}

The input request contains: 
\begin{itemize}
    \item instruction: a task described in natural language.
    \item image: a top-down photo of clothes, annotated with semantic keypoints labeled as P[i] (where i is the index).
    \item semantic keypoint dictionary: a dictionary mapping each P[i] to its language description.\\
\end{itemize}

You should determine whether the task is completed and the sequence of subtasks based on annotated keypoints. Each subtask includes the selected basic skill and contact points. \\

The response should be a dictionary in JSON form as follows:\\
\{
  ``completed": false,
  ``subtasks": [
    ``grasp(left sleeve)",
    ``moveto(center)",
    ``release()",
    ``grasp(right sleeve)",
    ``moveto(center)",
    ``release()",
    ``grasp(left shoulder, right shoulder)",
    ``moveto(left hem, right hem)",
    ``release()"
  ]
\}\\

I will provide some examples to demonstrate how to use these basic skills and the response format. 
\end{tcolorbox}
\end{center}

\newpage

\section{Additional Clothes Details}

This section presents the parameters of the clothes used in our experiment. The clothes cover various categories, shapes, textures, sizes, and materials. Each clothes is assigned a unique ID, and its size and material are also listed in the table. The size information indicates the width and height of the clothes when fully flat. In addition, we capture an RGB image of each clothes from a top-down view.
\vspace{0.8em}

\noindent
\begin{minipage}{0.30\linewidth}
  \centering
  \begin{tabular}{@{}ll@{}}
  \toprule
  No. & 1 \\ \midrule
  Size & 47 $\times$ 40 cm \\
  Material & \begin{tabular}[t]{@{}l@{}}Cotton \& Polyester\end{tabular} \\
  \bottomrule
  \end{tabular}
  \includegraphics[width=0.9\linewidth]{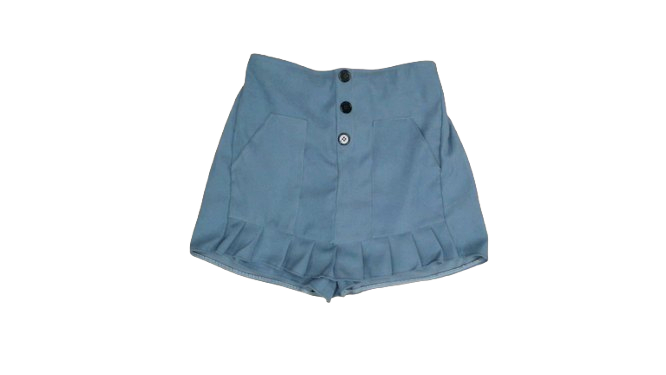}
\end{minipage}
\hfill
\begin{minipage}{0.30\linewidth}
  \centering
  \begin{tabular}{@{}ll@{}}
  \toprule
  No. & 2 \\ \midrule
  Size & 55 $\times$ 50 cm \\
  Material & \begin{tabular}[t]{@{}l@{}}Polyester\end{tabular} \\
  \bottomrule
  \end{tabular}
  \includegraphics[width=0.9\linewidth]{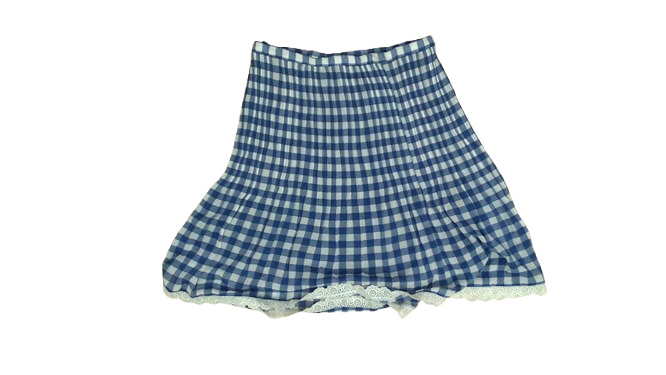}
\end{minipage}
\hfill
\begin{minipage}{0.30\linewidth}
  \centering
  \begin{tabular}{@{}ll@{}}
  \toprule
  No. & 3 \\ \midrule
  Size & 60 $\times$ 43 cm \\
  Material & \begin{tabular}[t]{@{}l@{}}Cotton \& Polyester\end{tabular} \\
  \bottomrule
  \end{tabular}
  \includegraphics[width=0.9\linewidth]{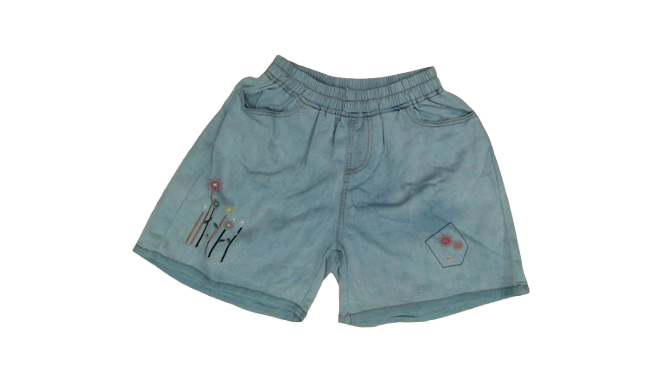}
\end{minipage}
\vspace{0.8em}

\noindent
\begin{minipage}{0.30\linewidth}
  \centering
  \begin{tabular}{@{}ll@{}}
  \toprule
  No. & 4 \\ \midrule
  Size & 40 $\times$ 30 cm \\
  Material & \begin{tabular}[t]{@{}l@{}}Cotton\end{tabular} \\
  \bottomrule
  \end{tabular}
  \includegraphics[width=0.9\linewidth]{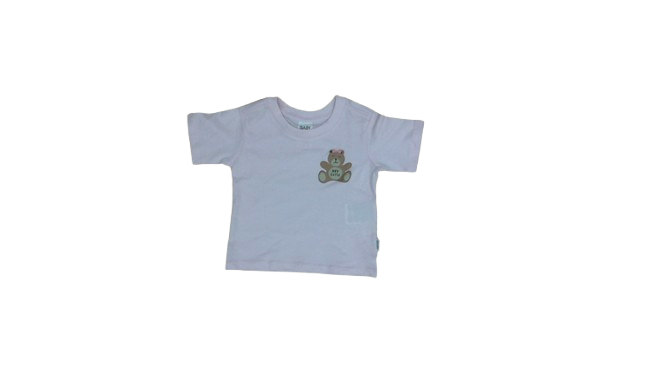}
\end{minipage}
\hfill
\begin{minipage}{0.30\linewidth}
  \centering
  \begin{tabular}{@{}ll@{}}
  \toprule
  No. & 5 \\ \midrule
  Size & 112 $\times$ 58 cm \\
  Material & \begin{tabular}[t]{@{}l@{}}Wool\end{tabular} \\
  \bottomrule
  \end{tabular}
  \includegraphics[width=0.9\linewidth]{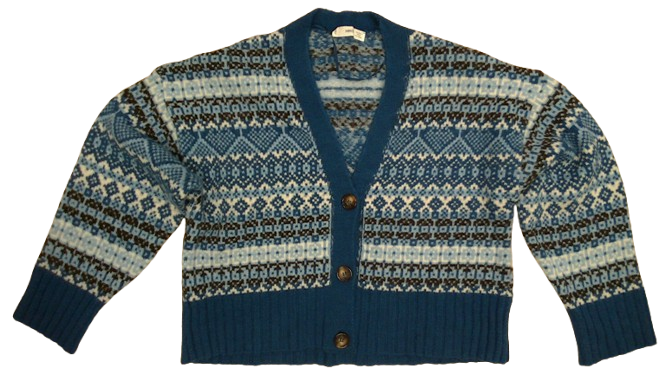}
\end{minipage}
\hfill
\begin{minipage}{0.30\linewidth}
  \centering
  \begin{tabular}{@{}ll@{}}
  \toprule
  No. & 6 \\ \midrule
  Size & 53 $\times$ 42 cm \\
  Material & \begin{tabular}[t]{@{}l@{}}Polyester \& Spandex\end{tabular} \\
  \bottomrule
  \end{tabular}
  \includegraphics[width=0.9\linewidth]{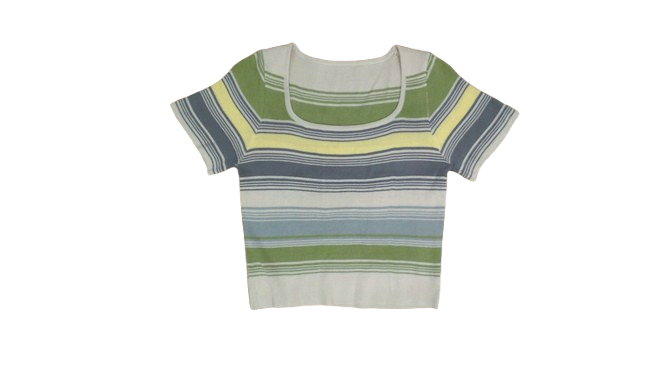}
\end{minipage}
\vspace{0.8em}

\noindent
\begin{minipage}{0.30\linewidth}
  \centering
  \begin{tabular}{@{}ll@{}}
  \toprule
  No. & 7 \\ \midrule
  Size & 46 $\times$ 35 cm \\
  Material & \begin{tabular}[t]{@{}l@{}}Cotton \& Polyester\end{tabular} \\
  \bottomrule
  \end{tabular}
  \includegraphics[width=0.9\linewidth]{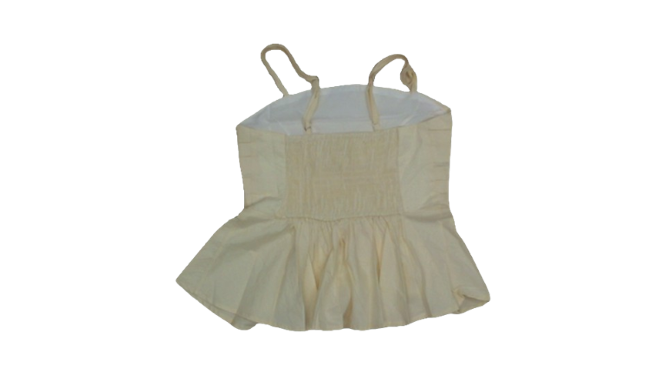}
\end{minipage}
\hfill
\begin{minipage}{0.30\linewidth}
  \centering
  \begin{tabular}{@{}ll@{}}
  \toprule
  No. & 8 \\ \midrule
  Size & 53 $\times$ 41 cm \\
  Material & \begin{tabular}[t]{@{}l@{}}Cotton \& Spandex\end{tabular} \\
  \bottomrule
  \end{tabular}
  \includegraphics[width=0.9\linewidth]{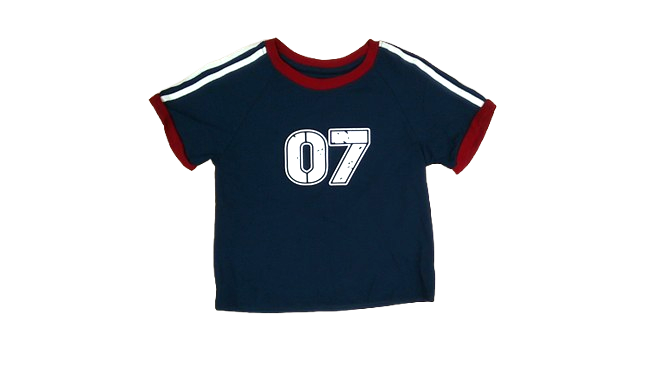}
\end{minipage}
\hfill
\begin{minipage}{0.30\linewidth}
  \centering
  \begin{tabular}{@{}ll@{}}
  \toprule
  No. & 9 \\ \midrule
  Size & 18 $\times$ 30 cm \\
  Material & \begin{tabular}[t]{@{}l@{}}Cotton\end{tabular} \\
  \bottomrule
  \end{tabular}
  \includegraphics[width=0.9\linewidth]{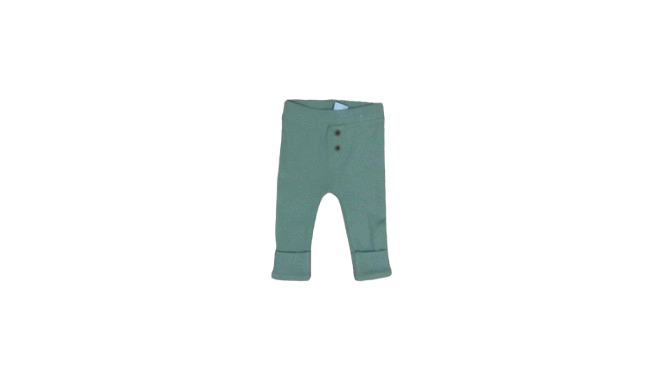}
\end{minipage}
\vspace{0.8em}

\noindent
\begin{minipage}{0.30\linewidth}
  \centering
  \begin{tabular}{@{}ll@{}}
  \toprule
  No. & 10 \\ \midrule
  Size & 53 $\times$ 41 cm \\
  Material & \begin{tabular}[t]{@{}l@{}}Nylon\end{tabular} \\
  \bottomrule
  \end{tabular}
  \includegraphics[width=0.9\linewidth]{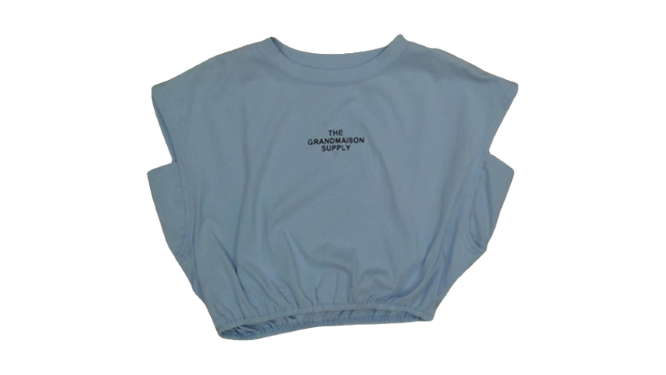}
\end{minipage}
\hfill
\begin{minipage}{0.30\linewidth}
  \centering
  \begin{tabular}{@{}ll@{}}
  \toprule
  No. & 11 \\ \midrule
  Size & 72 $\times$ 51 cm \\
  Material & \begin{tabular}[t]{@{}l@{}}Polyester \& Spandex\end{tabular} \\
  \bottomrule
  \end{tabular}
  \includegraphics[width=0.9\linewidth]{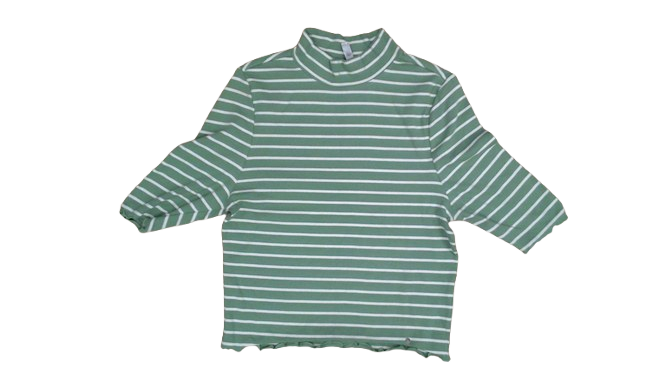}
\end{minipage}
\hfill
\begin{minipage}{0.30\linewidth}
  \centering
  \begin{tabular}{@{}ll@{}}
  \toprule
  No. & 12 \\ \midrule
  Size & 46 $\times$ 57 cm \\
  Material & \begin{tabular}[t]{@{}l@{}}Cotton \& Spandex\end{tabular} \\
  \bottomrule
  \end{tabular}
  \includegraphics[width=0.9\linewidth]{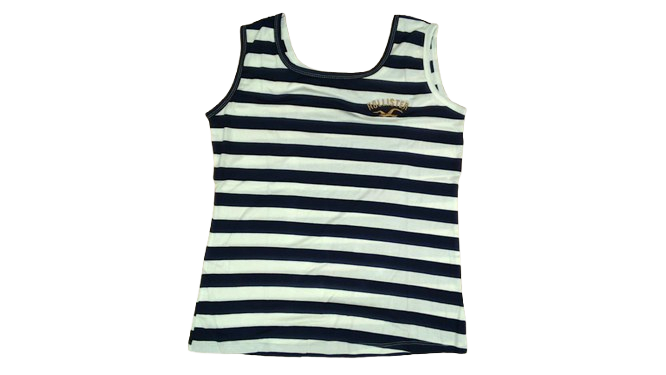}
\end{minipage}
\vspace{0.8em}

\noindent
\begin{minipage}{0.30\linewidth}
  \centering
  \begin{tabular}{@{}ll@{}}
  \toprule
  No. & 13 \\ \midrule
  Size & 70 $\times$ 40 cm \\
  Material & \begin{tabular}[t]{@{}l@{}}Polyester\end{tabular} \\
  \bottomrule
  \end{tabular}
  \includegraphics[width=0.9\linewidth]{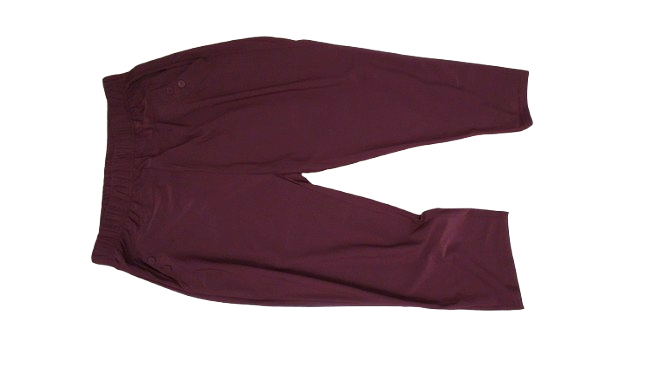}
\end{minipage}
\hfill
\begin{minipage}{0.30\linewidth}
  \centering
  \begin{tabular}{@{}ll@{}}
  \toprule
  No. & 14 \\ \midrule
  Size & 101 $\times$ 42 cm \\
  Material & \begin{tabular}[t]{@{}l@{}}Silk\end{tabular} \\
  \bottomrule
  \end{tabular}
  \includegraphics[width=0.9\linewidth]{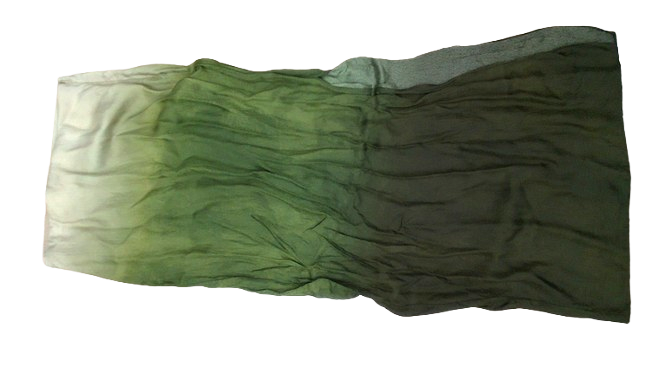}
\end{minipage}
\hfill
\begin{minipage}{0.30\linewidth}
  \centering
  \begin{tabular}{@{}ll@{}}
  \toprule
  No. & 15 \\ \midrule
  Size & 98 $\times$ 35 cm \\
  Material & \begin{tabular}[t]{@{}l@{}}Viscose\end{tabular} \\
  \bottomrule
  \end{tabular}
  \includegraphics[width=0.9\linewidth]{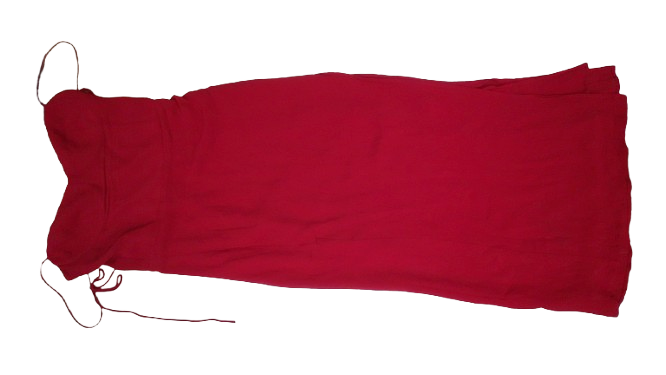}
\end{minipage}
\vspace{0.8em}

\end{document}